\documentclass[letterpaper]{article} 
\usepackage[preprint]{aaai2027}  
\usepackage[hyphens]{url}  
\usepackage{graphicx} 
\usepackage{natbib}  
\usepackage{caption} 
\usepackage{booktabs}
\usepackage{amsmath}
\usepackage{algorithm}
\usepackage{algorithmic}

\title{RECAP: Relation Evidence Calibration for Detecting Spatial Relation Hallucinations in Vision-Language Models}
\author{
    Feixiang Liu\textsuperscript{\rm 1,2},
    Qiang Qiu\textsuperscript{\rm 1},
    Qingyang Li\textsuperscript{\rm 1,2},
    Hui Xu\textsuperscript{\rm 1},
    Xueqi Cheng\textsuperscript{\rm 1}
}
\affiliations{
    \textsuperscript{\rm 1}State Key Laboratory of AI Safety, Institute of Computing Technology, CAS\\
    \textsuperscript{\rm 2}University of Chinese Academy of Sciences\\
    \{liufeixiang23s@mails.ucas.ac.cn, qiuqiang@ict.ac.cn, liqingyang24s@ict.ac.cn, xuhui@ict.ac.cn, cxq@ict.ac.cn\}
}

\begin{document}

\maketitle

\begin{abstract}
Vision-language models can answer spatial relation questions confidently even when the image supports an incompatible relation. We formulate relation-grounded selective prediction: accept or reject an already-produced yes/no answer by auditing its visual support, rather than treating uncertainty as evidence. RECAP, our relation-evidence calibration framework, compares image-conditioned likelihoods for a claim, its semantic contradictions, and optional one-sided supports, then converts these witnesses into an answer-conditioned rejection risk. A calibration-only gate preserves confidence as a veto when confidence is demonstrably informative and otherwise deploys relation evidence alone. Across 20 group/image-disjoint splits, RECAP lowers H-FPR@80 over confidence by between 2.0 and 17.9 points on VSR and raises Acc@80 by 3.0, 8.6, and 12.6 points on What'sUp for Qwen3-VL-8B, InternVL3.5-8B, and LLaVA-1.5-7B. It outperforms matched VCD-style visual contrast on all four primary metrics in all six settings. Full-pool VSR fallback, target-ranked GSR-Bench transfer, equal-budget supervised controls, and two additional checkpoints show a consistent operating principle: structured counterevidence complements certainty when confidence is misaligned, while the gate retains confidence when it is already useful.
\end{abstract}

\noindent\textbf{Code:} \url{https://github.com/SouthWinter/RECAP}

\section{Introduction}

Multimodal large language models (MLLMs) remain unreliable on spatial relations despite fluent object and scene descriptions. Broad benchmarks establish the need for structural grounding \citep{antol2015vqa,goyal2017vqav2,johnson2017clevr,hudson2019gqa,suhr2019nlvr2}, and relation-focused evaluations show that this weakness persists in strong models \citep{liu2023vsr,kamath2023whatsup,rajabi2024gsrbench,fu2024blink}. Both objects may be visible while the predicted relation is inverted or contradicted, sometimes with high confidence.

Existing signals target different failures: confidence measures certainty \citep{geifman2017selective,geifman2019selectivenet}, learned probes use internal states \citep{kogilathota2026halp}, and visual contrast measures perturbation sensitivity \citep{leng2024vcd,huang2024opera}. SpatialTrust learns from detector geometry \citep{imran2026spatialtrust}, whereas BCEA acquires claim-specific evidence with conformal control \citep{xu2026bcea}. RECAP instead audits a frozen yes/no answer by testing both the queried relation and its semantic contradiction, with answer-specific sufficient witnesses.

The difficult cases are those in which these signals disagree: a model can be certain and stable under generic visual perturbations even when a semantically incompatible relation is better supported. Conversely, low confidence may reflect scene difficulty rather than hallucination. A useful auditor must therefore test relation-specific alternatives and interpret them according to the answer being audited.

We formulate \emph{relation-grounded selective prediction}: accept or abstain from an existing spatial judgment rather than regenerate it. Spatial semantics supply structured counterevidence: \textit{left of} competes with \textit{right of}, \textit{above} with \textit{below}, and inverse forms such as \textit{inside}/\textit{contains} can be canonicalized under object order. An auditor can therefore compare a claim with contradictory witnesses while separating non-exclusive support relations.

This framing imposes two requirements. Evidence acquisition should hold the question and frozen answer fixed while changing only the relation being tested. Deployment should also avoid discarding confidence universally: confidence can remain informative for some models, so calibration rather than a post-hoc oracle must determine whether it complements relation evidence.

RECAP operationalizes this idea with a compact claim--contradiction graph probed through the same frozen MLLM. The score asks whether the fixed answer has a sufficient visual witness: claim support must overcome counterevidence for ``yes,'' whereas missing claim support or a supported contradiction can justify ``no.'' This answer-conditioned construction requires neither a learned detector nor model fine-tuning. Deployment pre-specifies two modes---relation evidence alone or the same evidence with a confidence veto---and a small group-disjoint calibration split selects the mode and threshold without held-out outcomes.

Across 20 group/image-disjoint splits, RECAP outperforms confidence and matched VCD-style contrast on VSR and What'sUp for all three headline models. The gate retains confidence for Qwen3-VL-8B but selects evidence-only RECAP for InternVL3.5-8B and LLaVA-1.5-7B when confidence is misaligned. Target-ranked GSR-Bench transfer, equal-budget controls, two additional checkpoints, and channel ablations show that semantic counterevidence adds information beyond certainty and pixel sensitivity.

\noindent In summary, our main contributions are threefold:
\begin{itemize}
\item We formulate relation-grounded selective prediction, a fixed-answer verification problem that separates model certainty from relation-specific visual support.
\item We introduce RECAP, which probes claim--contradiction graphs and converts image-conditioned relation evidence into an answer-conditioned rejection risk, with optional text normalization and a calibration-only gate for retaining useful confidence.
\item We establish the framework across five checkpoints using group-disjoint calibration, matched controls, external transfer, coverage-wide statistics, ablations, robustness tests, and cost analysis.
\end{itemize}

\section{Related Work}

\paragraph{Spatial Relation Benchmarks and Compositional Diagnostics.}
Scene graphs, compositional reasoning, and adversarial relation recognition study structural visual understanding \citep{krishna2017visualgenome,johnson2017clevr,hudson2019gqa,yang2019spatialsense}. Recent diagnostics show that vision-language models may match objects while ignoring word order or relational binding \citep{thrush2022winoground,parcalabescu2022valse,yuksekgonul2023bagofwords,ma2023crepe,hsieh2023sugarcrepe}. VSR, What'sUp, and GSR-Bench measure this failure through natural-image verification, controlled perturbations, and grounded spatial queries \citep{liu2023vsr,kamath2023whatsup,rajabi2024gsrbench}. RECAP instead asks whether one produced relation answer has enough image-grounded evidence to be accepted.

\paragraph{Multimodal Hallucination Detection and Relation Errors.}
Object-hallucination methods and benchmarks detect unsupported entities or coarse answer inconsistency \citep{rohrbach2018objecthallucination,li2023pope,liu2023lrvinstruction,guan2024hallusionbench,wang2023amber}, whereas relation errors can occur when both objects are perceived correctly. MMRel and Reefknot expose object--relation inconsistencies \citep{nie2024mmrel,zheng2025reefknot}. HALP learns architecture-specific probes over hidden representations for pre-generation detection \citep{kogilathota2026halp}; RECAP uses only output likelihoods and fits no detector, while converting relation evidence into a selective risk for one fixed answer.

\paragraph{Contrastive and Post-hoc Hallucination Control.}
VCD contrasts original and perturbed images, OPERA modifies decoding, and Woodpecker rewrites outputs \citep{leng2024vcd,huang2024opera,yin2023woodpecker}. Pelican and TACO decompose free-form responses before verification or refinement \citep{sahu2024pelican,liu2026taco}; constraint-aware prompting regenerates spatial answers \citep{wu2025constraintaware}. RECAP preserves the answer and tests it against incompatible relation witnesses. Matched VCD and constraint-aware answer-change controls isolate this distinction.

\paragraph{Selective Prediction and Multimodal Uncertainty.}
Selective prediction trades coverage for risk; calibration and conformal methods quantify uncertainty or control prediction sets \citep{chow1970reject,geifman2017selective,guo2017calibration,angelopoulos2023conformal}. Multimodal variants use question neighborhoods, follow-up evidence, or perturbation consistency \citep{khan2024consistency,srinivasan2024selective,zhang2024vluncertainty}. SpatialTrust learns from detector boxes, overlap, and detection quality \citep{imran2026spatialtrust}; those privileged geometric features are unavailable in our no-grounding-annotation protocol. Most closely, BCEA shows that a horizontal flip supplies evidence for left/right claims and folds acquisition into conformal calibration \citep{xu2026bcea}. Its score decides whether to assert a relation claim; our compatible experiment adapts that flip score to the relation asserted by a frozen yes/no answer. RECAP differs by explicitly competing claim and contradiction likelihoods across the evaluated relation graph and by conditioning the rejection rule on the answer. We compare the compatible score under the same group-disjoint selective protocol rather than claim BCEA's separate finite-sample certificate.

\section{Method: RECAP}

\subsection{Problem Setup and Overview}

Each example contains an image $I$, object mentions $(o_s,o_o)$, relation $r$, and a yes/no query. A frozen MLLM produces $\hat{y}\in\{\mathrm{yes},\mathrm{no}\}$; RECAP leaves it unchanged and assigns risk $R$ for selective acceptance.

For any yes/no probe query $q$, we compute the likelihood margin
\begin{equation}
m(q) = \log p(\mathrm{yes}\mid q) - \log p(\mathrm{no}\mid q),
\end{equation}
implemented as the difference between no and yes negative log-likelihoods. The standard confidence baseline uses only the direct-answer support,
\begin{equation}
R_{\mathrm{conf}} = - |m(q_{\mathrm{orig}})|.
\end{equation}
All risks are oriented so that higher values indicate higher rejection risk after ranking.

\begin{figure*}[t]
\centering
\includegraphics[width=\textwidth]{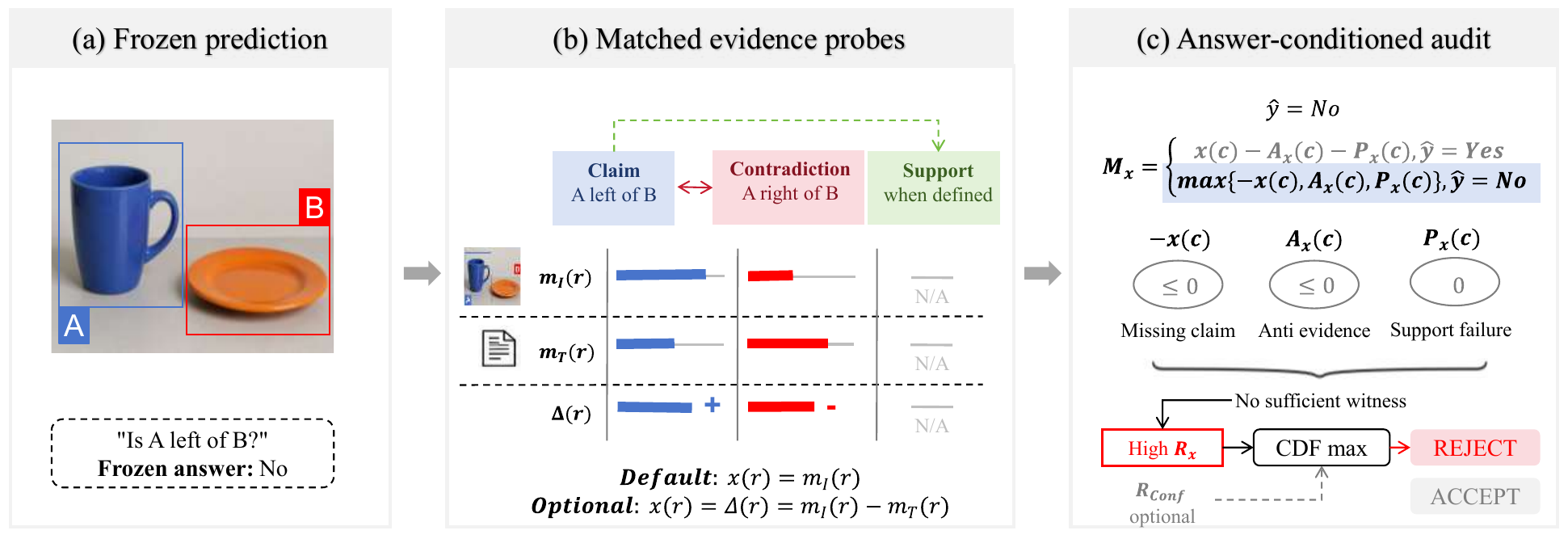}
\caption{Overview of RECAP. Given a frozen relation answer, RECAP probes the claim, contradictions, and supports when defined, then applies answer-conditioned auditing. The default channel uses $x=m_I$ for RECAP-Img; optional text-reference normalization uses $x=\Delta=m_I-m_T$ for RECAP-Cal. When calibration retains confidence as a veto, CDF max combines empirical calibration-CDF values rather than raw risk scales.}
\label{fig:recap-overview}
\end{figure*}

Figure~\ref{fig:recap-overview} summarizes the pipeline: canonicalize the claim, probe its contradiction/support graph on the image, optionally subtract matched text references, and aggregate relation-specific rejection risk. The supplement gives the complete risk construction.

\subsection{Relation Graph Construction}

RECAP represents local relation semantics as a directed constraint graph $G=(V,E_A,E_S,\beta)$. This graph is a falsification schema rather than a prompt list: its edges specify which visual witnesses can contradict a positive claim and which one-sided consequences can expose missing support. Nodes $V$ are canonical binary relation predicates over an ordered object pair. For a queried claim $c\in V$, outgoing contradiction edges define $A(c)=\{a:(c,a)\in E_A\}$, and support edges define $S(c)=\{s:(c,s)\in E_S\}$ with fixed weights $\beta_s$. Contradictory witnesses conflict with a positive claim, but they are not exhaustive complements for every negative answer; support witnesses encode one-sided checks that should usually hold when the claim is true.

\begin{table}[t]
\centering
{\small
\setlength{\tabcolsep}{4pt}
\begin{tabular}{llll}
\toprule
Family & Claim $c$ & $A(c)$ & $S(c)$ \\
\midrule
Horizontal & \textit{left\_of} & \textit{right\_of} & -- \\
Vertical & \textit{above} & \textit{below} & -- \\
Depth & \textit{in\_front\_of} & \textit{behind} & -- \\
Topology & \textit{inside} & \textit{contains} & -- \\
Vertical/contact & \textit{on} & \textit{below} & \textit{above} \\
\bottomrule
\end{tabular}
}
\caption{RECAP relation graph after canonicalizing reverse predicates by swapping object roles. Support is a soft one-sided witness.}
\label{tab:relation-graph}
\end{table}

Role-swapped aliases are canonicalized, e.g., \textit{contains}$(A,B)$ as \textit{inside}$(B,A)$. Non-exclusive relations remain supports rather than contradictions: for \textit{on}, \textit{above} is a weak one-sided witness.

Graph extension fixes three schema decisions before evaluation: role-swapped aliases, mutually exclusive witnesses for the same ordered pair/view, and optional one-sided supports with fixed weights. We omit ambiguous edges because spatial descriptions can co-occur. The reusable object is therefore a conservative constraint graph, not an enumeration of prompt variants.
This construction also distinguishes semantic coverage from edge reliability. Adding a relation node does not require connecting it to every neighboring predicate: an absent edge means that the method declines to treat the pair as reliable counterevidence. We retain the canonical depth contradiction as a candidate edge, but treat its viewpoint-dependent reliability as an empirical question rather than an unconditional logical guarantee. The supplement evaluates this design through family holdout, edge validation, and reliability gating.

\subsection{Probe Construction}

Each graph node becomes a yes/no probe using the original objects and the fixed template ``Is [subject] [relation] [object]?'' For relation $u$, the image-conditioned margin is $m_I(u)$. We also record a matched text-only margin $m_T(u)$ and define
\begin{equation}
\Delta(u) = m_I(u) - m_T(u).
\end{equation}
The default risk uses $m_I$ directly. The optional subtraction treats $m_T$ as a \emph{text-reference preference}, not a causal language prior: it may mix prompt/token bias, relation frequency, and object co-occurrence. Direct queries are reused only when their margins exactly match explicit claim probes.

\subsection{Evidence Aggregation}

For evidence channel $x(u)$, define contradiction and support summaries
\begin{equation}
A_x(c)=\max_{a\in A(c)}x(a),\qquad
P_x(c)=\sum_{s\in S(c)}\beta_s[-x(s)]_+,
\end{equation}
with $A_x(c)=0$ when $A(c)$ is empty, so a missing edge is neutral. The positive-claim evidence is
\begin{equation}
E_x(c)=x(c)-A_x(c)-P_x(c).
\end{equation}
RECAP-Img uses $x=m_I$ and is the default risk. RECAP-Cal substitutes $x=\Delta$ to test matched text-reference normalization under exactly the same graph and answer semantics. Weights are fixed before evaluation; only the soft \textit{on}$\rightarrow$\textit{above} witness uses $\beta_s=0.5$.

\subsection{Risk Scores and Deployment Modes}

The negation of a claim need not equal any single contradiction. RECAP therefore treats missing claim support, positive contradiction evidence, and failed support evidence as alternative witnesses for a predicted ``no'':
\begin{equation}
M_x=\begin{cases}
E_x(c), & \hat y=\mathrm{yes},\\
\max\{-x(c),A_x(c),P_x(c)\}, & \hat y=\mathrm{no},
\end{cases}
\end{equation}
\begin{equation}
R_x=-M_x.
\end{equation}
The maximum is a soft disjunction: any sufficient witness can support ``no'' without being canceled by another channel. We denote $R_{m_I}$ as RECAP-Img and $R_\Delta$ as RECAP-Cal. This matched construction directly tests whether text subtraction adds value beyond structured image evidence.

For deployment, let $\mathcal{C}$ be the group-disjoint calibration set and define the empirical calibration CDF of risk $z$ as
\begin{equation}
\widehat F_z^{\mathcal C}(t)=\frac{1}{|\mathcal C|}
\sum_{i\in\mathcal C}\mathbf{1}\{R_z^{(i)}\le t\}.
\end{equation}
The confidence-gated selector is
\begin{equation}
R_{\mathrm{sel}}^{\mathcal C}
=\max\!\left\{
\widehat F_{\mathrm{conf}}^{\mathcal C}(R_{\mathrm{conf}}),
\widehat F_{\mathrm{rel}}^{\mathcal C}(R_{\mathrm{rel}})
\right\}.
\end{equation}
It accepts an answer only when both global confidence and relation evidence are low-risk. A calibration quantile $\tau_\gamma$ targets coverage $\gamma$, and deployment accepts iff $R_{\mathrm{sel}}^{\mathcal C}\le\tau_\gamma$. We instantiate $R_{\mathrm{rel}}$ with RECAP-Img in the main table; no test-pool rank enters this score.
We call this combined risk RECAP-Selector.

Deployed RECAP has exactly two modes: RECAP-Img alone or RECAP-Selector. RECAP-Cal and single-channel scores are mechanism controls. RECAP fits no learned risk model: its graph, probes, and risk formula are fixed. Deployment nevertheless uses a labeled group-disjoint calibration split to set the acceptance threshold and test whether confidence is eligible to remain as a veto under the two conditions below. The resulting mode and threshold are frozen before utility is measured on held-out groups.
Specifically, 1,000 group-bootstrap resamples form Bonferroni-adjusted one-sided bounds with nominal joint 95\% coverage for direct H-FPR and confidence Hall-AUC. The veto is eligible only when the H-FPR upper bound is below $0.5$ and the Hall-AUC lower bound is above $0.5$. The first is a fixed majority-reliability reference---fewer than half of negative claims are falsely asserted---whereas the second is chance ranking. These interpretable boundaries are not a risk guarantee, but they avoid model-specific tuned cutoffs. Otherwise deployment conservatively uses RECAP-Img.

\subsection{Probe Cost and Reuse}

Smaller probe subsets compute the same risks through the exact claim-reuse rule above; the supplement verifies sample- and metric-level equivalence.

\section{Experiments}

We test whether relation evidence improves selective prediction beyond confidence and VCD-style contrast, which channels explain the gains, and whether conclusions transfer across models, domains, prompts, relation families, and probe budgets.

\begin{table*}[t]
\centering
{\small
\setlength{\tabcolsep}{2pt}
\begin{tabular}{llrrrrrrrr}
\toprule
 & & \multicolumn{4}{c}{VSR} & \multicolumn{4}{c}{What'sUp} \\
\cmidrule(lr){3-6}\cmidrule(lr){7-10}
Model & Risk & Acc@80$\uparrow$ & H-FPR@80$\downarrow$ & Err-AUC$\uparrow$ & Hall-AUC$\uparrow$ & Acc@80$\uparrow$ & H-FPR@80$\downarrow$ & Err-AUC$\uparrow$ & Hall-AUC$\uparrow$ \\
\midrule
Qwen3-VL-8B & Confidence & \textbf{85.6} & 16.8 & 74.1 & 69.6 & 82.7 & 23.6 & 69.5 & 68.8 \\
 & VCD-Contrast & 84.6 & 20.7 & 68.5 & 58.5 & 82.6 & 24.1 & 73.9 & 72.8 \\
 & RECAP$^\dagger$ & 85.3 & \textbf{14.8} & \textbf{74.3} & \textbf{72.9} & \textbf{85.7} & \textbf{19.3} & \textbf{80.1} & \textbf{79.6} \\
\midrule
InternVL3.5-8B & Confidence & 62.1 & 94.6 & 62.1 & 60.3 & 42.8 & 82.8 & 67.0 & 66.8 \\
 & VCD-Contrast & 61.5 & 84.3 & 57.5 & 58.2 & 49.7 & 71.2 & 85.6 & 85.7 \\
 & RECAP$^\dagger$ & \textbf{66.8} & \textbf{80.2} & \textbf{73.0} & \textbf{74.7} & \textbf{51.4} & \textbf{70.4} & \textbf{94.1} & \textbf{94.2} \\
\midrule
LLaVA-1.5-7B & Confidence & 64.4 & 72.5 & 57.6 & 52.3 & 46.0 & 75.8 & 52.3 & 51.1 \\
 & VCD-Contrast & 67.0 & 63.8 & 64.2 & 62.8 & 55.3 & 60.7 & 78.5 & 77.9 \\
 & RECAP$^\dagger$ & \textbf{70.2} & \textbf{54.6} & \textbf{71.6} & \textbf{77.1} & \textbf{58.6} & \textbf{58.0} & \textbf{90.7} & \textbf{90.9} \\
\bottomrule
\end{tabular}
}
\caption{Held-out means over 20 group-disjoint splits. Each 20\% calibration split fixes the 80\% coverage threshold; Deployed RECAP ($\dagger$) chooses between RECAP-Img and RECAP-Selector using calibration only. Bold marks the best risk in each setting. All entries are percentages.}
\label{tab:main-results}
\end{table*}

\subsection{Setup}

\paragraph{Data and models.}
The primary natural/controlled benchmarks are the pre-specified graph-covered VSR random-test subset and controlled What'sUp \citep{liu2023vsr,kamath2023whatsup}; an image-disjoint COCO/GQA two-object GSR-Bench subset tests external transfer \citep{rajabi2024gsrbench}. We evaluate Qwen3-VL-8B, InternVL3.5-8B, and LLaVA-1.5-7B \citep{liu2023llava,chen2024internvl,bai2025qwen3vl}, plus Qwen3-VL-2B as a scale stress test and LLaVA-OneVision-7B as a newer-architecture check \citep{li2024llavaonevision}. Compact tables abbreviate the headline models as Qwen3-8B, InternVL-8B, and LLaVA-7B.

\paragraph{Metrics and comparisons.}
We report selective Acc@80, H-FPR@80, error AUROC, and hallucination AUROC, and use direct FPR/FNR to diagnose answer bias. H-FPR is the predicted-yes fraction among accepted negatives; Hall-AUC uses false-positive relation assertions as positives, whereas Err-AUC uses either wrong answer. One global threshold determines the reported 80\% coverage for both labels; class-specific coverage checks appear in the supplement. We use 80\% as a moderate-abstention point; 70\%/90\% and AURC appear in the supplement. Comparisons share one frozen likelihood interface: confidence, Prompt-SC, clean--perturbed VCD, an answer-aware confidence model, and a supervised raw-probe model fitted on the same labeled calibration groups. Additional controls isolate each relation channel.

\paragraph{Protocol.}
We retain 1,249 of 2,195 VSR test statements for graph probing using only their normalized relation family; a full-pool deployment falls back to confidence for the 946 relations outside the graph (supplement). Controlled What'sUp contains 3,280 statements in 820 choice groups, and image-disjoint GSR contains 1,438 statements in 719 groups. The yes/no NLL argmin is the direct answer; probes use the same full-continuation margin. Formulas and graph edges are fixed across models and datasets. We report means over 20 group/image-disjoint 20\% calibration splits; thresholds and the RECAP mode are fixed on calibration before held-out evaluation. Clustered bootstrap resamples whole choice groups or images.

\subsection{Main Results}

Table~\ref{tab:main-results} reports the deployment protocol rather than pooled test-set ranking. RECAP improves all four metrics over VCD in every model--dataset setting. Against confidence, it lowers VSR H-FPR@80 by 2.0, 14.4, and 17.9 points for Qwen, InternVL, and LLaVA, respectively; on What'sUp it raises Acc@80 by 3.0, 8.6, and 12.6 points and Hall-AUC by 10.8--39.8 points. Semantic counterevidence therefore supplies more useful rejection evidence than generic pixel sensitivity.

The calibration gate exposes two repeatable regimes. It retains the confidence veto in 19/20 Qwen VSR and 20/20 Qwen What'sUp splits, but selects evidence-only RECAP in every InternVL/LLaVA split. Thus confidence is preserved when it already ranks hallucinations and replaced only when calibration establishes misalignment. Realized coverage remains near 80\%; no held-out outcome selects the mode or threshold. Paired clustered intervals and pooled common-coverage diagnostics appear in the supplement.

The same policy extends to all 2,195 VSR statements by using confidence outside the relation graph. Relative to confidence on the full pool, this fallback lowers H-FPR@80 by 1.4/8.1/9.7 points and raises Hall-AUC by 2.3/8.0/13.4 points for Qwen/InternVL/LLaVA, respectively.

\begin{table*}[t]
\centering
{\small
\setlength{\tabcolsep}{3pt}
\begin{tabular}{llrrrrrr}
\toprule
& & \multicolumn{2}{c}{Conf-AnsLogit} & \multicolumn{2}{c}{Probe-Logit} & \multicolumn{2}{c}{RECAP-Img} \\
\cmidrule(lr){3-4}\cmidrule(lr){5-6}\cmidrule(lr){7-8}
Model & Data & Acc@80 & H-FPR & Acc@80 & H-FPR & Acc@80 & H-FPR \\
\midrule
Qwen3-VL-8B & VSR & \textbf{85.1} & \textbf{12.5} & 81.1 & 18.4 & 85.0 & 17.4 \\
 & What'sUp & 94.0 & 6.3 & \textbf{95.0} & \textbf{5.0} & 85.1 & 20.4 \\
InternVL3.5-8B & VSR & 64.8 & 81.2 & 65.3 & 80.9 & \textbf{66.7} & \textbf{80.2} \\
 & What'sUp & 50.5 & 70.9 & 51.1 & 70.7 & \textbf{51.2} & \textbf{70.6} \\
LLaVA-1.5-7B & VSR & 66.3 & 58.8 & 68.7 & 58.8 & \textbf{69.9} & \textbf{55.2} \\
 & What'sUp & 55.6 & 59.6 & 56.9 & 58.9 & \textbf{58.6} & \textbf{58.0} \\
\bottomrule
\end{tabular}
}
\caption{Equal-budget supervised controls over 20 group-disjoint splits. Conf-AnsLogit uses confidence and answer identity; Probe-Logit fits error from the unaggregated direct and relation-probe margins; RECAP-Img keeps its risk fixed. Every method uses the same 20\% labeled calibration groups. Full AUC results appear in the supplement.}
\label{tab:fair-confidence}
\end{table*}

These controls map the two operating regimes more sharply. Supervision exploits a strong in-domain answer prior on Qwen/What'sUp, where Probe-Logit is best. Without fitting a risk model, RECAP-Img has better Acc@80 and H-FPR@80 than Probe-Logit in the other five settings and wins all four metrics over Conf-AnsLogit in the four InternVL/LLaVA settings. The fixed structured prior is therefore most useful where confidence and answer identity are insufficient, while learned fusion remains attractive when representative labels expose stable domain bias.

\subsection{Generalization and Mechanism Checks}

\paragraph{Newer-architecture validation.}

\begin{table}[t]
\centering
{\small
\setlength{\tabcolsep}{3pt}
\begin{tabular}{llrrr}
\toprule
Data & Risk & Acc@80 & H-FPR@80 & Hall-AUC \\
\midrule
VSR & Confidence & 68.9 & 80.8 & 69.1 \\
 & VCD & 73.8 & 62.1 & 78.2 \\
 & RECAP & \textbf{74.9} & \textbf{58.5} & \textbf{80.9} \\
\midrule
WU & Confidence & 56.5 & 62.0 & 68.0 \\
 & VCD & 63.5 & 52.2 & 81.8 \\
 & RECAP & \textbf{65.3} & \textbf{49.2} & \textbf{84.4} \\
\bottomrule
\end{tabular}
}
\caption{Held-out LLaVA-OneVision-7B deployment over 20 group/image-disjoint splits. Calibration fixes each threshold and selects RECAP-Img in all splits. Entries are percentages.}
\label{tab:onevision-main}
\end{table}

On OneVision, deployed RECAP lowers H-FPR@80 by 22.3 points on VSR and 12.8 on What'sUp while raising Hall-AUC by 11.8 and 16.4 points. It also improves all three metrics over VCD. The eligibility test rejects the confidence veto in all 20 splits on both datasets; pooled RECAP-Cal diagnostics remain in the supplement.

\paragraph{Answer-bias controls.}
Before rejection, Qwen3-VL-8B has lower yes rate (44.5--53.3) and direct FPR (21.8--27.4) than InternVL/LLaVA (yes rate 75.7--92.5; FPR 67.2--86.3). A pure yes penalty cannot rank within a fixed direct-answer subset. On What'sUp, RECAP-Img improves Err-AUC within both direct-yes and direct-no subsets for all three models (Table~\ref{tab:answer-bias-main}); in the pooled exact-coverage diagnostic, FPR falls while FNR remains at most 4.0. Thus the fixed relation risk is not merely a global direct-answer penalty.

\begin{table}[t]
\centering
{\small
\setlength{\tabcolsep}{2pt}
\begin{tabular}{lrrrr}
\toprule
& \multicolumn{2}{c}{Conf.$\to$policy} & \multicolumn{2}{c}{Conf.$\to$RECAP-Img} \\
\cmidrule(lr){2-3}\cmidrule(lr){4-5}
Model & FPR@80 & FNR@80 & Yes AUC & No AUC \\
\midrule
Qwen3-8B & 23.6$\to$19.4 & 2.4$\to$2.8 & 91.6$\to$93.2 & 66.9$\to$80.1 \\
InternVL-8B & 82.8$\to$70.6 & 0.1$\to$0.5 & 83.5$\to$90.6 & 67.2$\to$73.1 \\
LLaVA-7B & 75.6$\to$57.9 & 1.8$\to$4.0 & 73.4$\to$85.2 & 65.2$\to$85.4 \\
\bottomrule
\end{tabular}
}
\caption{Offline answer-bias controls on What'sUp at pooled 80\% coverage. Deployed RECAP uses RECAP-Selector for Qwen and RECAP-Img otherwise; the last two columns condition on the fixed direct answer.}
\label{tab:answer-bias-main}
\end{table}

\paragraph{Relation families.}
RECAP is strongest for visually exclusive contradictions. On LLaVA VSR topology, RECAP-Img raises Hall-AUC from 59.1 to 89.0 and lowers H-FPR@80 from 22.5 to 9.5; left/right and vertical also benefit. Leave-one-family-out transfer improves H-FPR for both confidence-misaligned models across all families (Table~\ref{tab:family-transfer-main}). Depth remains the boundary case for Qwen because viewpoint, occlusion, and scale weaken both probes. A calibrated depth fallback restores Qwen but lowers ranking quality for other models, so we report it as a reliability diagnostic rather than replace the fixed deployment rule post hoc.

\subsection{Evidence-Regime Analysis}

\begin{figure}[t]
\centering
\includegraphics[width=\linewidth]{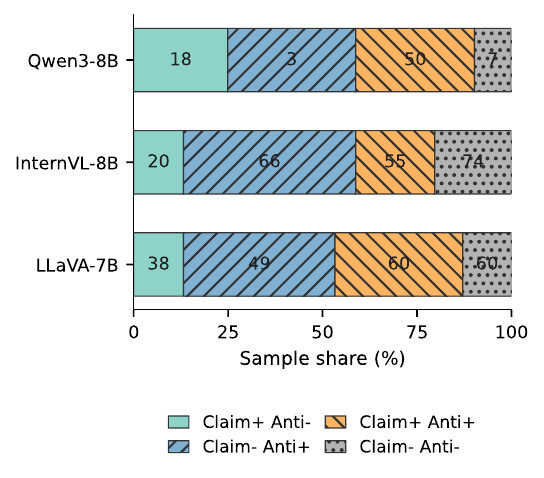}
\caption{What'sUp RECAP-Cal evidence regimes defined by the signs of $\Delta$. Widths show sample share; numbers show direct-answer error rates.}
\label{fig:evidence-quadrants-main}
\end{figure}

To test whether RECAP tracks the direct-answer prior, we partition What'sUp by the signs of the text-referenced claim and strongest anti-relation lifts $\Delta$ (Figure~\ref{fig:evidence-quadrants-main}). The four states represent clean claim evidence, clean counterevidence, relation inconsistency, and weak visual evidence. Widths show prevalence; labels show direct-answer error.

\begin{table}[t]
\centering
{\small
\setlength{\tabcolsep}{2pt}
\begin{tabular}{llrr}
\toprule
Model & Family & Conf. C/F & RECAP C/F \\
\midrule
Qwen3-8B & Left/right & 89.9/24.0 & 81.8/12.3 \\
 & Vertical & 81.2/31.5 & 85.0/30.5 \\
 & Depth & 52.2/11.1 & 67.0/15.4 \\
\midrule
InternVL-8B & Left/right & 60.1/93.9 & 73.7/71.2 \\
 & Vertical & 93.3/57.1 & 93.2/51.2 \\
 & Depth & 83.7/99.4 & 72.9/93.1 \\
\midrule
LLaVA-7B & Left/right & 83.4/64.1 & 80.2/44.1 \\
 & Vertical & 79.1/85.4 & 93.8/75.7 \\
 & Depth & 72.9/89.2 & 58.1/63.9 \\
\bottomrule
\end{tabular}
}
\caption{Leave-one-family-out transfer on What'sUp. C/F is coverage/H-FPR using the mode and threshold selected on the other families.}
\label{tab:family-transfer-main}
\end{table}

Qwen's clean regimes have only 18\% and 3\% error, consistent with its selected confidence veto. InternVL reaches 55--74\% error in the jointly high/low regimes, and LLaVA reaches 60\% in both. Thus identical answer directions can carry different relation-evidence risk, explaining both RECAP's gains and its correlated failures rather than merely relabeling global yes bias.

\subsection{External Transfer: GSR-Bench}

After source-qualified image-ID comparison against both primary benchmarks, we remove 9 shared COCO images (24 statements), leaving 544 images, 719 groups, and 1,438 GSR-Bench statements \citep{rajabi2024gsrbench}. RECAP receives RGB and the query, not boxes, masks, or depth.

Using only pooled primary-domain calibration, the eligibility test selects RECAP-Selector for Qwen and RECAP-Img for InternVL/LLaVA in all 20 source splits before GSR-Bench inspection. At a common target-ranked 80\% coverage, these modes lower H-FPR by 1.9--24.0 points and raise Hall-AUC by 13.9--37.7 points (Table~\ref{tab:gsrbench-summary}). This establishes target-domain ranking transfer: unlabeled target scores set the 80\% quantile, but no target labels select the mode. The supplement separately transfers source CDFs and thresholds unchanged, exposing the resulting coverage drift.

\begin{table}[t]
\centering
{\small
\setlength{\tabcolsep}{3pt}
\begin{tabular}{lccc}
\toprule
Model & Acc@80 & H-FPR@80 & Hall-AUC \\
\midrule
Qwen3-8B & 91.5$\to$91.2 & 2.5$\to$0.6 & 81.0$\to$94.9 \\
InternVL-8B & 67.3$\to$75.8 & 47.1$\to$23.1 & 54.1$\to$91.8 \\
LLaVA-7B & 56.8$\to$66.1 & 94.6$\to$76.9 & 64.2$\to$93.0 \\
\bottomrule
\end{tabular}
}
\caption{External GSR-Bench ranking at common target-ranked 80\% coverage, shown as Confidence$\to$the source-selected image-evidence mode. Target scores fix coverage; no GSR labels or annotations select the mode.}
\label{tab:gsrbench-summary}
\end{table}

\subsection{Task Extension: Multi-Choice Relation Selection}

Controlled What'sUp also permits a four-way diagnostic over the same image--object pair. Direct image probes reach 93.5/65.0/56.5 accuracy for Qwen, InternVL, and LLaVA; on 646 calibration-disjoint groups, the fixed graph raises Qwen from 93.7 to 95.7 and InternVL from 64.9 to 85.0, while LLaVA changes little. This is evidence that the probes contain relation-selection information, not part of the binary selective-auditing claim; full MRR, family, top-$k$, and edge analyses remain in the supplement.

\subsection{Ablations, Cost, and Failure Behavior}

\paragraph{Ablations and cost.}
Matched Prompt-SC, VCD, and constraint-aware answer-change controls separate semantic counterevidence from prompt instability and pixel sensitivity (supplement). ImgPair, Anti-Delta, and CalPair identify contradiction as the reusable channel, while RECAP-Img versus RECAP-Cal shows that text-reference subtraction is optional and model-dependent. Confidence uses two likelihood calls, Prompt-SC ten, and optimized RECAP-Img averages 4.8--5.0; a confidence-first staged policy reduces this to about 3.2 calls (supplement). Conf-AnsLogit and a richer supervised combiner quantify what can be gained by fitting an in-domain error separator rather than using a fixed risk.

\begin{table}[t]
\centering
{\small
\setlength{\tabcolsep}{2.2pt}
\begin{tabular}{lrrrrrr}
\toprule
Image risk & Q-V & Q-W & I-V & I-W & L-V & L-W \\
\midrule
Confidence & 69.9 & 68.9 & 60.3 & 66.9 & 52.4 & 51.0 \\
Claim only & 72.3 & 67.3 & 64.5 & 74.5 & 55.2 & 61.2 \\
Symmetric pair & 61.1 & 78.0 & 74.6 & 94.3 & 74.8 & 90.6 \\
AC claim+anti & 68.0 & 79.1 & 74.8 & 94.2 & 77.2 & 90.9 \\
AC + support & 68.7 & 79.3 & 74.9 & 94.2 & 77.3 & 90.9 \\
\bottomrule
\end{tabular}
}
\caption{Core mechanism ablation by Hall-AUC. Q/I/L denote Qwen, InternVL, and LLaVA; V/W denote VSR and What'sUp. AC is answer conditioning.}
\label{tab:mechanism-main}
\end{table}

Table~\ref{tab:mechanism-main} holds the image channel fixed. Claim-only evidence is insufficient; claim--anti competition produces the largest gain for confidence-misaligned models, and answer conditioning consistently improves the symmetric pair except for the near-tied InternVL/What'sUp case. The single soft support edge adds little overall, as expected from its narrow coverage. The confidence veto is isolated in Table~\ref{tab:main-results}; prompt robustness and recurrent self-audit failure regimes appear in the supplement.

\paragraph{Statistical reliability.}
Paired clustered bootstrap resamples whole images or choice groups and compares each pre-specified RECAP mode with confidence at common 80\% coverage. Table~\ref{tab:bootstrap-main} shows that every H-FPR and Hall-AUC interval excludes zero. On Qwen/VSR, where Acc@80 and Err-AUC are statistically tied, the safety-oriented gains remain positive; all four metrics improve significantly in the other five settings. Error AURC confirms that the result persists across coverage: the policy ties confidence on Qwen/VSR and improves the other five settings (full curves and intervals in the supplement).

\section{Discussion and Limitations}

The central result is a division of labor between certainty and evidence. Confidence is effective when its ranking is aligned with relation errors, as for Qwen; answer-conditioned claim--contradiction evidence is substantially stronger when that alignment fails, as for InternVL and LLaVA. The calibration-only gate turns this observation into one deployment rule rather than a post-hoc model choice, retaining confidence in the former regime and relation evidence in the latter.

Contradiction is the main reusable channel: it tests a claim against a semantic alternative that generic uncertainty and visual degradation do not express. Answer conditioning then distinguishes evidence required for ``yes'' from sufficient witnesses for ``no.'' The graph serves as a conservative interface for these tests; support and text-reference channels are useful diagnostics but are not necessary for the primary gains. External transfer and relation-family holdout further indicate that the resulting ordering is not confined to one benchmark split.

The decomposed risk also provides an audit trail: a rejected answer can be traced to missing claim evidence, a stronger incompatible relation, or failed support rather than only to an opaque confidence scalar. These components identify the semantic test responsible for abstention and make relation-level failure patterns directly comparable across models, prompts, and relation families.

The present evaluation isolates fixed-answer verification for ordered object pairs and a validated relation vocabulary. Extending it to free-form multi-object outputs requires relation extraction, reference resolution, and measured edge reliability. Same-model probes can inherit correlated perception errors, likelihood access is required, and RECAP-Img averages 4.8--5.0 calls. Jointly positive claim/anti evidence is a detectable inconsistency regime; a competitive no-branch penalty improves some rankings but harms fixed-coverage utility on smaller/newer checkpoints, so the supplement reports it without post-hoc replacement. The risk formula itself is fixed, while a small labeled calibration split sets coverage and selects the confidence veto. Accordingly, RECAP is a selective auditor rather than an answer-correction system or safety certificate; model, domain, graph, or prompt changes require recalibration.

\begin{table}[!t]
\centering
{\small
\setlength{\tabcolsep}{3pt}
\begin{tabular}{lrr}
\toprule
Setting & $\Delta$H-FPR@80 & $\Delta$Hall-AUC \\
\midrule
Qwen / VSR & +2.2 [0.1, 4.0] & +3.5 [0.6, 6.8] \\
Qwen / WU & +4.2 [3.0, 5.6] & +10.9 [9.1, 12.5] \\
InternVL / VSR & +14.5 [11.2, 18.3] & +14.7 [10.7, 18.4] \\
InternVL / WU & +12.2 [10.9, 13.4] & +27.4 [25.6, 29.0] \\
LLaVA / VSR & +17.4 [14.5, 20.7] & +24.8 [21.0, 28.8] \\
LLaVA / WU & +17.7 [16.4, 19.1] & +39.9 [37.9, 41.5] \\
\bottomrule
\end{tabular}
\vspace{5pt}

\begin{tabular}{lrr}
\toprule
 & \multicolumn{2}{c}{Error AURC $\downarrow$} \\
\cmidrule(lr){2-3}
Setting & Confidence & Deployed RECAP \\
\midrule
Qwen / VSR & 9.3 & 9.3 \\
Qwen / WU & 11.5 & 8.1 \\
InternVL / VSR & 31.9 & 24.9 \\
InternVL / WU & 43.0 & 25.8 \\
LLaVA / VSR & 30.2 & 22.7 \\
LLaVA / WU & 44.8 & 22.3 \\
\bottomrule
\end{tabular}
}
\caption{Statistical reliability and coverage robustness. Top: paired clustered-bootstrap gains over confidence with 95\% intervals; positive values favor RECAP. Bottom: error AURC across the full coverage range. All values are percentages.}
\label{tab:bootstrap-main}
\end{table}

\section{Conclusion}

RECAP audits fixed spatial-relation answers by comparing a frozen claim with semantic contradictions and conditioning rejection evidence on the answer. Across five checkpoints, relation evidence improves hallucination ranking; a calibration-only gate retains confidence when informative and otherwise uses evidence alone. The controls support this division of labor between certainty and visual support.

\bibliography{aaai2027}

\clearpage
\section*{Supplementary Material}
\makeatletter
\setlength{\@fptop}{0pt}
\setlength{\@fpsep}{10pt}
\setlength{\@fpbot}{0pt plus 1fil}
\setlength{\@dblfptop}{0pt}
\setlength{\@dblfpsep}{10pt}
\setlength{\@dblfpbot}{0pt plus 1fil}
\makeatother
\renewcommand{\topfraction}{0.95}
\renewcommand{\bottomfraction}{0.90}
\renewcommand{\textfraction}{0.05}
\renewcommand{\floatpagefraction}{0.70}
\renewcommand{\dbltopfraction}{0.95}
\renewcommand{\dblfloatpagefraction}{0.70}
\setcounter{topnumber}{4}
\setcounter{bottomnumber}{2}
\setcounter{totalnumber}{6}
\setcounter{dbltopnumber}{3}
\setlength{\textfloatsep}{12pt plus 2pt minus 2pt}
\setlength{\floatsep}{10pt plus 2pt minus 2pt}
\setlength{\dbltextfloatsep}{12pt plus 2pt minus 2pt}
\setlength{\dblfloatsep}{10pt plus 2pt minus 2pt}

\section{Experimental Protocol Details}

\paragraph{Evaluation setting.}
All experiments use frozen multimodal models as post-hoc relation verifiers. RECAP does not train a detector or risk model, fine-tune a model, change the model's direct answer, or use privileged grounding annotations at test time. Its deployment does use labeled calibration groups to set coverage and determine whether confidence is eligible as a veto. For each sample, the input to the auditor is the RGB image, a binary spatial-relation statement, the model's direct yes/no answer, and likelihood scores from follow-up yes/no probes. The primary benchmarks are VSR and the controlled What'sUp subset; the external GSR-Bench evaluation uses the COCO-Spatial-Two and GQA-Spatial-Two splits after image-ID de-duplication against both primary benchmarks. Although GSR-Bench provides boxes, masks, and depth annotations, RECAP uses only the image and the binary relation query in that transfer experiment.

\paragraph{Sample construction.}
The official VSR random test split contains 2,195 statements. Before any model inference, we retain the four relation families with pre-specified graph constraints, yielding 1,249 statements: 245 left/right, 543 vertical (above/below/on), 169 topology (inside/contains), and 292 depth (front/behind). The remaining 946 statements comprise 225 proximity, 64 interaction, and 657 other relations. Selection uses only normalized relation identity and is independent of labels, predictions, and risk scores. Our graph analyses use the 1,249 covered statements; Section~\ref{sec:full-vsr-fallback} evaluates an operational confidence fallback on the complete 2,195-statement pool. Relation phrases are matched longest-first at word boundaries before terminal copulas and leading articles are removed. What'sUp controlled A/B yields 3,280 statements (1,640 left/right, 824 vertical, 816 depth) in 820 choice groups. GSR-Bench initially has 1,462 statements over 553 images. Normalizing dataset-qualified COCO/GQA image IDs, including zero padding, reveals 9 COCO images shared with VSR and none with controlled What'sUp. Removing their 24 statements leaves 1,438 statements (1,070 left/right, 320 vertical, 48 depth), 719 groups, and 544 images.

\paragraph{Task scope.}
The evaluation isolates relation-level auditing: each example contains one ordered object pair and a binary query over a validated relation vocabulary. This controlled setting separates evidence ranking from relation extraction, reference resolution, multi-object reasoning, and answer correction. The graph-extension protocol below specifies how new relations can be admitted conservatively; all headline numbers use the fixed graph in Table~\ref{tab:supp-relation-graph}.

\paragraph{Models and frozen likelihood interface.}
We evaluate Qwen3-VL-8B-Instruct, InternVL3.5-8B-HF, and LLaVA-1.5-7B-HF; Qwen3-VL-2B-Instruct and LLaVA-OneVision-7B provide scale and architecture checks. Each checkpoint uses its released preprocessing and chat template. Image prompts include the native image item; text references omit it but keep the same question, template, and assistant continuation. With one leading-space delimiter, $m=\mathrm{NLL}(\text{no})-\mathrm{NLL}(\text{yes})$ over full continuations, $\hat y=\mathrm{yes}$ iff $m\geq0$, and confidence risk is $-|m|$. RECAP formulas are fixed; labeled deployment calibration only sets thresholds or chooses between two pre-specified modes.

\paragraph{Inference and resampling details.}
Inference uses one NVIDIA A800 80GB GPU. OneVision uses bfloat16 with scaled dot-product attention, while LLaVA-1.5 uses float16. Selective metrics use stable ascending risk order, with ties following canonical dataset order. Bootstrap intervals use 1000 clustered resamples with seed 13 and report the point estimate plus/minus the half-width of the 95\% percentile interval. The sampling unit is the controlled choice group when available and otherwise the image ID. For split seed $k\in\{0,\ldots,19\}$, a unit enters calibration when the first eight hexadecimal digits of the MD5 digest of the concatenated seed and unit identifier, interpreted as an integer, are divisible by five; all remaining units are held out. Thus all correlated options or image queries remain in the same partition.

\paragraph{Metrics and operating point.}
At 80\% coverage we report Acc@80 and H-FPR@80, plus error and hallucination AUROC. Let $z_{\mathrm{hall}}=\mathbf{1}[\hat y=\mathrm{yes},y=\mathrm{no}]$ and $z_{\mathrm{err}}=\mathbf{1}[\hat y\neq y]$. H-FPR is predicted-yes frequency among accepted negatives; Hall-AUC and Err-AUC rank risk against $z_{\mathrm{hall}}$ and $z_{\mathrm{err}}$, respectively. The highest-risk 20\% are rejected. Bootstrap, coverage curves, and relation-family results quantify stability.

\paragraph{Prompt and risk selection protocol.}
Deployment pre-specifies RECAP-Img and RECAP-Selector; RECAP-Cal and other channel variants remain controls. External transfer freezes the mode from pooled primary calibration. Prompt-SC uses the original query plus four paraphrases: ``According to the image, [query]?'', ``Look at the image carefully. [Query]?'', ``Answer yes or no using only visual evidence: [query]?'', and ``Based on what is visible, is it true that [query]?''. Prompt robustness instead shares four wrappers across the direct query and all probes: unmodified, ``According to the image, answer this question:'', ``Look carefully at the image before answering:'', and ``Using only visible evidence, answer yes or no:''. It averages templates and rank correlations without best-prompt selection.

\paragraph{Online thresholding protocol.}
The selector is deployed without test-set ranks by storing empirical calibration CDFs for $R_{\mathrm{conf}}$ and the selected relation-evidence risk. Each new score is mapped to its calibration percentile, the selector takes the maximum percentile, and a calibration quantile fixes the target-coverage threshold. Mode eligibility uses the fixed-reference joint test defined in the main paper, not held-out performance. Its H-FPR boundary requires a majority of negative claims not to be falsely asserted; only the Hall-AUC boundary is a chance comparison. At joint confidence levels of 90\%, 95\%, and 99\%, the test selects the veto in 19, 19, and 17 of 20 Qwen VSR splits and all 20 Qwen What'sUp splits, but never for InternVL3.5-8B or LLaVA-1.5-7B. The main paper reports held-out threshold realization across all 20 group-disjoint splits.

\paragraph{Comparison methods.}
All fixed-score baselines use the same yes/no likelihood interface and require no risk-model training. Confidence is the direct likelihood-margin risk. Direct binary entropy is monotone in the absolute yes/no margin and therefore induces the same ranking. Prompt-SC evaluates the original query and four fixed semantic paraphrases, keeps the original answer, and uses the weakest answer-aligned margin as risk. VCD-Contrast compares clean-image probes with deterministic blur/downsample/grayscale perturbed-image probes for the same claim and contradiction relations. Claim-Delta tests claim-only image-minus-text lift, TextPair tests text-reference relation competition, ImgPair tests image-side competition, Anti-Delta isolates contradiction lift, and CalPair applies text subtraction without support penalties. Conf-AnsLogit fits error from confidence, answer identity, and their interaction. Probe-Logit instead receives the unaggregated direct, claim, contradiction, and support margins plus relation-family indicators. Both learned controls use exactly the same labeled calibration groups as RECAP's threshold calibration; the richer Logit-Hall combiner is retained as an upper-bound diagnostic.

\section{Probe and Relation-Graph Details}

\paragraph{Probe families.}
Each RECAP probe is a yes/no statement of the form ``Is [subject] [relation] [object]?'' The default RECAP-Img risk evaluates the claim, contradiction relations, and optional support relations on the original image. RECAP-Cal additionally evaluates exactly matched text-only references for the same statements. Transformed-image, flip, mask, and crop probes are diagnostics rather than inputs to the default risk. The cost analysis verifies that RECAP-Img needs only the graph relations referenced by its fixed formula.

\paragraph{Relation graph.}
Table~\ref{tab:supp-relation-graph} lists the relation graph used to construct contradiction and support probes. The graph is compact and precision-oriented: it covers the evaluated relation families while admitting only validated semantic constraints. The ``canonicalized form'' column indicates role-swapped aliases used to reduce duplicate probe logic; for example, a query about the object being right of the subject can be represented as the subject being left of the object after swapping roles.

To extend the graph, a new relation should be added only after assigning its role in the same constraint schema: canonical aliases, mutually exclusive witnesses, optional support witnesses, and support weights. Ambiguous or co-occurring relations should be left without contradiction edges unless visual exclusivity is clear. This reliability-aware extension rule prioritizes high-precision counterevidence and makes each admitted edge auditable.

\paragraph{Graph extension protocol.}
RECAP uses a relation-constraint schema rather than selecting a new scoring formula for each relation. A candidate relation is first normalized into aliases and role-swapped canonical forms. Candidate contradiction edges may be proposed from lexical opposites, dataset answer options, or an external lexicon, but an edge is admitted only when the two predicates are visually incompatible for the same ordered object pair under the benchmark's reference frame. Candidate support edges are admitted only as one-sided witnesses: they may be expected when the claim is true, but their failure is used as a soft penalty rather than a logical proof of falsehood. \textit{Near}/\textit{far} is excluded from the evaluated graph because its exclusivity depends on scale and context; viewpoint-dependent depth and co-occurring contact/vertical descriptions likewise require conservative edges. This protocol makes graph extension auditable; later experiments evaluate semi-automatic candidate generation and held-out edge validation under the same admission principle.

\begin{algorithm}[t]
\caption{RECAP-Img risk construction with optional text calibration}
\label{alg:supp-recap}
\begin{algorithmic}[1]
\REQUIRE Evaluation pool $\mathcal{D}$ with images, relation queries, and direct answers.
\ENSURE RECAP-Img risk and, when requested, RECAP-Cal risk for each example.
\FOR{each example $(I,q_{\mathrm{orig}},o_s,o_o,r,\hat y)$ in $\mathcal{D}$}
\STATE Canonicalize $r$ and the ordered object pair into a claim relation $c$.
\STATE Retrieve contradictory witnesses $A(c)$ and support witnesses $S(c)$.
\FOR{each relation $u\in\{c\}\cup A(c)\cup S(c)$}
\STATE Build a yes/no relation probe for $u(o_s,o_o)$.
\STATE Compute image-conditioned margin $m_I(u)$.
\STATE Set $x(u)=m_I(u)$ for RECAP-Img.
\STATE Optionally compute $m_T(u)$ and set $x(u)=m_I(u)-m_T(u)$ for RECAP-Cal.
\ENDFOR
\STATE Set $A_x(c)=\max_{a\in A(c)}x(a)$ and
$P_x(c)=\sum_{s\in S(c)}\beta_s\max(0,-x(s))$.
\STATE Set $E_x(c)=x(c)-A_x(c)-P_x(c)$.
\IF{$\hat y=\mathrm{yes}$}
\STATE Set answer-conditioned margin $M_x=E_x(c)$.
\ELSE
\STATE Set $M_x=\max\{-x(c),A_x(c),P_x(c)\}$.
\ENDIF
\STATE Set relation risk $R_x=-M_x$.
\ENDFOR
\STATE Return $R_{m_I}$ (and optional $R_\Delta$); deployment maps $R_{\mathrm{conf}}$ and $R_{m_I}$ through calibration CDFs before taking their maximum.
\end{algorithmic}
\end{algorithm}

For a calibration set $\mathcal C$, deployment maps each component through
$\widehat F_z^{\mathcal C}(t)=|\mathcal C|^{-1}
\sum_{i\in\mathcal C}\mathbf{1}\{R_z^{(i)}\le t\}$ and uses
$R_{\mathrm{sel}}^{\mathcal C}=\max\{
\widehat F_{\mathrm{conf}}^{\mathcal C}(R_{\mathrm{conf}}),
\widehat F_{\mathrm{rel}}^{\mathcal C}(R_{\mathrm{rel}})\}$.
The calibration $\gamma$-quantile fixes the acceptance threshold. Offline diagnostic tables may rank risks within a fixed evaluation pool, but deployment never uses test-pool ranks.

\begin{table*}[t]
\centering
{\small
\begin{tabular}{llll}
\toprule
Relation family & Claim relation & Contradiction relation(s) & Support / canonicalized form \\
\midrule
Left/right & left\_of & right\_of & right\_of canonicalizes to left\_of with swapped roles \\
Left/right & right\_of & left\_of & left\_of canonical form with swapped roles \\
Vertical & above & below & below canonicalizes to above with swapped roles \\
Vertical & below & above & above canonical form with swapped roles \\
Vertical & on & below & above is a soft support relation ($\beta=0.5$) \\
Topology & inside & contains & contains canonicalizes to inside with swapped roles \\
Topology & contains & inside & inside canonical form with swapped roles \\
Depth & in\_front\_of & behind & behind canonicalizes to in\_front\_of with swapped roles \\
Depth & behind & in\_front\_of & in\_front\_of canonical form with swapped roles \\
\bottomrule
\end{tabular}
}
\caption{RECAP relation graph used for probe construction. Support relations are soft witnesses, not logical requirements.}
\label{tab:supp-relation-graph}
\end{table*}

\section{Answer-Conditioned Evidence}

The symmetric risk conversion multiplies a relation-pair margin by the sign of the direct answer. For a predicted ``no,'' this can be read too strongly: an anti-relation is a witness against the claim, but it is not generally the exhaustive meaning of \emph{not claim}. For example, \emph{not on} does not imply \emph{below}. RECAP instead interprets the same graph evidence conditionally on the frozen answer. For either $x=m_I$ (RECAP-Img) or $x=m_I-m_T$ (RECAP-Cal), define
\begin{equation}
M_x(c,\hat y)=
\begin{cases}
E_x(c), & \hat y=\mathrm{yes},\\
\max\{-x(c),A_x(c),P_x(c)\},
& \hat y=\mathrm{no},
\end{cases}
\end{equation}
where
\begin{align}
E_x(c)&=x(c)-A_x(c)-P_x(c),\\
P_x(c)&=\sum_{s\in S(c)} \beta_s \max(0,-x(s)).
\end{align}
We set $A_x(c)=0$ when $A(c)$ is empty, making an absent contradiction edge neutral.
Thus missing claim support can justify ``no'' even when no anti-relation is strongly supported; anti evidence remains an optional sufficient witness rather than a required logical complement. The risk is $R_x=-M_x$. The uniform six-setting mechanism table in the main paper shows that claim-only evidence is insufficient, claim--anti competition supplies most of the reusable gain, and answer conditioning improves the symmetric pair in five settings while remaining nearly tied in the sixth. The text-calibration analysis below then isolates the optional $m_T$ subtraction.

\paragraph{Jointly positive claim and anti evidence.}
A predicted ``no'' can receive a strong anti witness even when the claim margin is also positive. We therefore tested a competitive alternative that replaces $A_x(c)$ in the no branch by $A_x(c)-x(c)$, leaving predicted-yes scores unchanged. Table~\ref{tab:competitive-no} reports held-out changes over 20 splits. The alternative modestly improves ranking for Qwen but worsens fixed-coverage accuracy and H-FPR for Qwen3-VL-2B and OneVision; it is nearly inert for InternVL/LLaVA because their jointly positive no cases already lie well inside the accepted region. Across both datasets, such cases comprise 0.7--5.2\% of samples across checkpoints. The original rule retains 44--100\% of them at pooled 80\% coverage, with retained error rates of 2.1--17.7\%. Thus this is a measurable inconsistency regime, but not a uniformly erroneous one: forcing claim--anti competition rejects useful negative answers in some settings. We retain the pre-specified soft disjunction and expose this subset rather than selecting the alternative post hoc.

\begin{table}[t]
\centering
{\scriptsize
\setlength{\tabcolsep}{1.2pt}
\begin{tabular}{lrrrr}
\toprule
Model & $\Delta$Acc & $\Delta$H-FPR & $\Delta$E-AUC & $\Delta$H-AUC \\
\midrule
Qwen3-8B & $+0.0/+0.2$ & $-0.1/-0.2$ & $-0.1/+2.5$ & $+1.4/+2.6$ \\
InternVL-8B & $+0.0/+0.0$ & $+0.0/+0.0$ & $-0.0/+0.0$ & $+0.1/+0.0$ \\
LLaVA-7B & $+0.0/+0.0$ & $+0.0/+0.0$ & $+0.0/+0.2$ & $-0.0/+0.1$ \\
Qwen3-2B & $-0.1/-1.2$ & $+0.3/+1.7$ & $+0.4/+0.6$ & $+0.5/+0.6$ \\
OneVision-7B & $-0.1/-0.6$ & $+0.2/+0.3$ & $+0.1/-0.5$ & $+0.2/-0.6$ \\
\bottomrule
\end{tabular}
}
\caption{Competitive predicted-no ablation. Values are percentage-point changes from the original soft disjunction over 20 held-out splits; V/W denotes VSR/What'sUp. Positive is better except for H-FPR.}
\label{tab:competitive-no}
\end{table}

\section{Comparison Baselines}

\paragraph{VCD-style visual contrast baseline.}
The VCD-style baseline uses the same clean-image likelihood interface as RECAP but contrasts it with deterministic visual degradation rather than semantic contradiction alone. The perturbed view downsamples each dimension by a factor of 16 with bilinear interpolation (minimum 8 pixels), restores the original size bicubically, applies Gaussian blur with radius $\max(2,\min(H,W)/48)$, and blends the result 25\% toward RGB gray $(127,127,127)$. VCD-Contrast compares clean and perturbed evidence for the claim and anti-relation. This construction makes VCD a strong fixed-score control for generic visual sensitivity while keeping the distinction from RECAP clear: VCD perturbs pixels, whereas RECAP structures relation evidence around claim--contradiction support.

\paragraph{BCEA horizontal-flip comparison.}
BCEA uses claim-specific interventions and conformal recalibration to decide whether a relation claim can be asserted. Its left/right component is directly compatible with our frozen likelihood interface. For the relation $r_{\hat y}$ asserted by the frozen answer---the query relation for ``yes'' and its left/right contradiction for ``no''---we adapt its score as $s_{\mathrm{flip}}=m_I(r_{\hat y})-m_{\mathrm{flip}(I)}(r_{\hat y})$ and use $-s_{\mathrm{flip}}$ as rejection risk. Table~\ref{tab:bcea-flip} evaluates this score on the left/right subsets with the same 20 group/image-disjoint calibration splits. BCEA-Flip improves Hall-AUC over confidence on What'sUp but not VSR; Deployed RECAP is tied with confidence on VSR and substantially stronger on What'sUp. This is a score-level interface comparison, not a reproduction of BCEA's separate acquisition policy or finite-sample certificate.

\begin{table}[t]
\centering
{\footnotesize
\setlength{\tabcolsep}{2.5pt}
\begin{tabular}{llrrrr}
\toprule
Data & Risk & Cov. & Acc. & H-FPR & Hall-AUC \\
\midrule
VSR & Confidence & 76.7 & 87.4 & \textbf{11.1} & 78.0 \\
 & BCEA-Flip & 77.3 & 85.5 & 20.8 & 69.3 \\
 & RECAP-Img & 77.8 & 85.4 & 19.0 & 65.9 \\
 & Deployed RECAP & 77.3 & \textbf{87.5} & 11.8 & \textbf{78.1} \\
\midrule
What'sUp & Confidence & 80.9 & 84.7 & 21.5 & 70.4 \\
 & BCEA-Flip & 79.4 & 83.8 & 23.3 & 76.9 \\
 & RECAP-Img & 80.0 & \textbf{92.6} & \textbf{10.4} & \textbf{90.6} \\
 & Deployed RECAP & 80.5 & 90.6 & 13.1 & 86.6 \\
\bottomrule
\end{tabular}
}
\caption{Compatible BCEA score comparison for Qwen3-VL-8B left/right claims. Calibration thresholds target 80\% coverage; Cov. reports held-out realization. Entries are 20-split held-out means in percent.}
\label{tab:bcea-flip}
\end{table}

\paragraph{Constraint-aware answer-change baseline.}
We adapt the constraint-aware spatial prompting baseline described in the main paper to the same selective-prediction interface. Bidirectional prompts label the queried objects A and B and ask the model to check B relative to A before A relative to B. Combined prompts additionally ask for a clearly identifiable reference object C and transitive A--C/B--C checks when such an object exists. Both variants keep the image and binary query fixed and return only a yes/no continuation. If $m_v$ is the yes/no margin under constrained variant $v$ and $\hat y$ is the frozen original answer, the continuous rejection risk is
\begin{equation}
R_{\mathrm{CA},v}=-s(\hat y,m_v),
\end{equation}
where $s(\mathrm{yes},m)=m$ and $s(\mathrm{no},m)=-m$. Thus weak constrained support for the original answer raises rejection risk; the binary sign-change indicator is retained only as a diagnostic. Combined is pre-specified as the headline variant, while Bidirectional measures the cheaper constraint. The headline risk needs two probes in total, one original and one constrained.

\begin{table*}[t]
\centering
{\footnotesize
\setlength{\tabcolsep}{4pt}
\begin{tabular}{llrrrr}
\toprule
Data & Risk & Acc@80 $\uparrow$ & H-FPR@80 $\downarrow$ & Err-AUC $\uparrow$ & Hall-AUC $\uparrow$ \\
\midrule
VSR & Confidence & \textbf{85.4} & 17.1 & \textbf{74.4} & 69.9 \\
 & Prompt-SC & 84.4 & 17.9 & 73.3 & 68.7 \\
 & VCD-Contrast & 84.5 & 20.5 & 68.6 & 58.4 \\
 & CA-Bidirectional & 82.0 & 29.7 & 67.9 & 49.2 \\
 & CA-Combined & 82.6 & 25.9 & 70.0 & 56.5 \\
 & RECAP-Img & 85.3 & \textbf{16.9} & 73.5 & 68.7 \\
\midrule
What'sUp & Confidence & 82.7 & 23.6 & 69.5 & 68.9 \\
 & Prompt-SC & 82.3 & 23.5 & 72.5 & 72.4 \\
 & VCD-Contrast & 82.6 & 24.1 & 73.9 & 72.8 \\
 & CA-Bidirectional & 77.8 & 31.2 & 49.0 & 46.6 \\
 & CA-Combined & 77.0 & 32.1 & 52.4 & 50.5 \\
 & RECAP-Img & \textbf{85.1} & \textbf{20.4} & \textbf{80.0} & \textbf{79.3} \\
\bottomrule
\end{tabular}
}
\caption{Constraint-aware prompting adapted as a selective rejection risk for Qwen3-VL-8B. CA uses constrained support for the fixed original answer rather than replacing that answer. All methods share the same frozen likelihood interface; bold marks the best risk per dataset and metric, and entries are percentages.}
\label{tab:constraint-aware-risk}
\end{table*}

Table~\ref{tab:constraint-aware-risk} separates answer improvement from error detection. CA-Combined changes 19.9\% of VSR and 12.7\% of What'sUp answers, but only 32.3\% and 18.3\% of those changes identify an originally wrong answer. It also shifts the yes rate from 53.3\% to 67.6\% on VSR and from 44.5\% to 53.5\% on What'sUp. Consequently, answer change mostly reflects prompt sensitivity and answer-direction bias rather than calibrated rejection evidence. The literal binary Combined-Flip risk is weaker still: its Err-AUC/Hall-AUC is 57.2/48.2 on VSR and 48.7/47.5 on What'sUp. In paired 1,000-draw group bootstrap, continuous CA-Combined loses 13.3 Hall-AUC points to Confidence on VSR (95\% interval $[-17.4,-9.2]$) and 18.4 on What'sUp ($[-20.0,-16.7]$). Its point Hall-AUC also trails RECAP-Img by 12.2 and 28.8 points. This control establishes that constraint-guided regeneration and selective acceptance require different evidence signals.

\section{Statistical and Deployment Validation}

We estimate uncertainty with a paired clustered bootstrap that resamples controlled choice groups for What'sUp and images for VSR. Confidence and the pre-specified RECAP mode use the same resampled units within each draw, so Table~\ref{tab:paired-bootstrap-gains} directly reports utility-oriented metric differences rather than comparing marginal intervals.

\begin{table*}[t]
\centering
{\footnotesize
\setlength{\tabcolsep}{3pt}
\begin{tabular}{lllrrrr}
\toprule
Model & Data & Mode & $\Delta$Acc@80 & $\Delta$H-FPR@80 & $\Delta$Err-AUC & $\Delta$Hall-AUC \\
\midrule
Qwen3-VL-8B & VSR & RECAP-Selector & +0.0 [-1.3,+0.9] & +2.2 [+0.1,+4.0] & +0.3 [-1.5,+2.4] & +3.5 [+0.6,+6.8] \\
 & What'sUp & RECAP-Selector & +2.8 [+2.0,+3.9] & +4.2 [+3.0,+5.6] & +10.7 [+9.0,+12.3] & +10.9 [+9.1,+12.5] \\
InternVL3.5-8B & VSR & RECAP-Img & +4.5 [+2.9,+6.5] & +14.5 [+11.2,+18.3] & +11.2 [+7.2,+15.1] & +14.7 [+10.7,+18.4] \\
 & What'sUp & RECAP-Img & +8.4 [+7.5,+9.3] & +12.2 [+10.9,+13.4] & +27.1 [+25.4,+28.8] & +27.4 [+25.6,+29.0] \\
LLaVA-1.5-7B & VSR & RECAP-Img & +5.7 [+3.8,+7.7] & +17.4 [+14.5,+20.7] & +14.1 [+10.3,+18.3] & +24.8 [+21.0,+28.8] \\
 & What'sUp & RECAP-Img & +12.7 [+11.7,+13.6] & +17.7 [+16.4,+19.1] & +38.5 [+36.5,+40.1] & +39.9 [+37.9,+41.5] \\
\bottomrule
\end{tabular}
}
\caption{Paired clustered-bootstrap utility gains over Confidence. H-FPR gains are Confidence minus RECAP; all others are RECAP minus Confidence, so positive values favor RECAP. Brackets are 95\% percentile intervals from 1000 choice-group/image resamples.}
\label{tab:paired-bootstrap-gains}
\end{table*}

The paired intervals sharpen the interpretation. Every H-FPR and Hall-AUC gain is positive at the 95\% level. On Qwen3-VL-8B VSR, Acc@80 and Err-AUC intervals include zero, while the safety-oriented H-FPR and Hall-AUC gains remain positive. All four gains are positive for the other five model--dataset settings, preserving the distinction between a complementary veto and an evidence-first ranker.

\paragraph{Prevalence-sensitive ranking.}
Hall-AUC treats false-positive assertions as positives but does not expose their base rate. Table~\ref{tab:prevalence-auprc} therefore reports hallucination prevalence and average precision (AP), whose random-ranking baseline equals prevalence. RECAP improves AP over confidence in all six settings, including the low-prevalence Qwen/VSR case and the high-prevalence InternVL/LLaVA settings.

\begin{table}[t]
\centering
{\footnotesize
\setlength{\tabcolsep}{2.5pt}
\begin{tabular}{llrrrr}
\toprule
Model & Data & Hall. $n$ & Prev. & Conf. AP & RECAP AP \\
\midrule
Qwen3-8B & VSR & 127 & 10.2 & 19.4 & 27.2 \\
 & What'sUp & 675 & 20.6 & 34.8 & 52.8 \\
InternVL-8B & VSR & 502 & 40.2 & 45.8 & 65.9 \\
 & What'sUp & 1928 & 58.8 & 66.7 & 95.5 \\
LLaVA-7B & VSR & 391 & 31.3 & 32.5 & 62.3 \\
 & What'sUp & 1694 & 51.6 & 50.3 & 91.5 \\
\bottomrule
\end{tabular}
}
\caption{Hallucination prevalence and average precision. A hallucination is a false-positive relation assertion; prevalence is its fraction among all statements. RECAP uses the pre-specified pooled policy mode for each model. Entries except counts are percentages.}
\label{tab:prevalence-auprc}
\end{table}

\paragraph{Eligibility-rule sensitivity.}
We also compare the fixed-reference rule with direct calibration-utility selection, which chooses RECAP-Img or RECAP-Selector by calibration H-FPR@80, breaking ties by accuracy and Hall-AUC. The two rules agree in every InternVL/LLaVA split and in 13/20 Qwen VSR and 19/20 Qwen What'sUp splits. Despite the seven different Qwen VSR choices, direct utility changes held-out Acc/H-FPR by only $-0.16/+0.13$ points there and $-0.07/+0.11$ on What'sUp relative to the fixed-reference gate. The interpretable criterion therefore achieves essentially the same held-out utility without selecting the rule by held-out outcomes.

\section{Text-Reference Calibration Effect}

Table~\ref{tab:text-calibration-effect} first isolates text subtraction in pairwise competition. ImgPair compares the image-conditioned claim margin against the strongest image-conditioned anti-relation margin; CalPair applies the same competition after subtracting matched text-reference margins. The effect is model-dependent: calibration helps Qwen VSR, is nearly neutral on the two Qwen/InternVL What'sUp settings, and favors image-only competition elsewhere. The answer-conditioned RECAP-Img result in the last column is stronger still on InternVL/LLaVA because it adds answer-specific witnesses and support semantics. We therefore treat $m_T(u)$ as an optional operational reference, not a causal language-prior estimate or a required component.

\begin{table*}[t]
\centering
{\footnotesize
\setlength{\tabcolsep}{3pt}
\begin{tabular}{llrrrrrr}
\toprule
Model & Data & ImgPair H-AUC & CalPair H-AUC & $\Delta$H-AUC & ImgPair Acc@80 & CalPair Acc@80 & RECAP-Img H-AUC \\
\midrule
Qwen3-VL-8B & VSR & 60.5 & 66.3 & +5.9 & 84.9 & 85.1 & 68.7 \\
Qwen3-VL-8B & What'sUp & 77.5 & 77.8 & +0.3 & 83.8 & 83.5 & 79.3 \\
InternVL3.5-8B & VSR & 74.5 & 72.2 & -2.3 & 66.3 & 65.2 & 74.9 \\
InternVL3.5-8B & What'sUp & 94.3 & 95.1 & +0.8 & 51.1 & 51.2 & 94.2 \\
LLaVA-1.5-7B & VSR & 74.6 & 71.8 & -2.8 & 69.9 & 68.9 & 77.3 \\
LLaVA-1.5-7B & What'sUp & 90.3 & 87.2 & -3.1 & 58.5 & 57.5 & 90.9 \\
\bottomrule
\end{tabular}
}
\caption{Effect of text-reference calibration. ImgPair compares claim and anti-relation image margins directly; CalPair subtracts matched text-reference margins. $\Delta$H-AUC is CalPair minus ImgPair. The final column reports RECAP-Img Hall-AUC. All values are percentages.}
\label{tab:text-calibration-effect}
\end{table*}

\section{Low-Cost Equivalence}

Table~\ref{tab:supp-probe-cost} reports the probe cost behind the cost summary in the main paper. For each model, dataset, and variant, we evaluated the required probe subset and compared it with the complete probe set. Across 36 model--dataset--variant checks, both the maximum metric difference and the maximum sample-risk difference were exactly zero; removing unused probes changes computation but not scores, rankings, or selection.
Table~\ref{tab:staged-cost} additionally simulates a cheaper online policy: confidence handles the easiest samples first, and RECAP is invoked only for the remaining uncertain samples. This staged variant is not used for the headline results; it quantifies the cost--utility trade-off available when latency is more important than maximizing relation-evidence ranking strength.

\begin{table}[t]
\centering
{\footnotesize
\setlength{\tabcolsep}{1.5pt}
\begin{tabular}{lrrrrrrr}
\toprule
Data & Conf. & P-SC & Anti & Img & Cal & Full & Selector \\
\midrule
VSR & 2.0 & 10.0 & 6.0 & 4.8 & 8.8 & 9.2 & 4.8 \\
What'sUp & 2.0 & 10.0 & 6.0 & 5.0 & 9.0 & 9.5 & 5.0 \\
\bottomrule
\end{tabular}
}
\caption{Average yes/no likelihood calls per sample. Img, Cal, and Selector denote RECAP-Img, RECAP-Cal, and RECAP-Selector.}
\label{tab:supp-probe-cost}
\end{table}

\begin{table}[t]
\centering
{\small
\setlength{\tabcolsep}{3pt}
\begin{tabular}{llrrr}
\toprule
Model & Data & Probes & Minutes & Sec./sample \\
\midrule
Qwen3-VL-8B & VSR & 10226 & 70.5 & 3.39 \\
Qwen3-VL-8B & What'sUp & 27064 & 182.1 & 3.33 \\
InternVL-8B & VSR & 10226 & 154.8 & 7.44 \\
InternVL-8B & What'sUp & 27064 & 327.4 & 5.99 \\
LLaVA-7B & VSR & 10226 & 48.3 & 2.32 \\
LLaVA-7B & What'sUp & 27064 & 140.8 & 2.58 \\
\bottomrule
\end{tabular}
}
\caption{Wall-clock time for evaluating the complete RECAP+VCD diagnostic probe set on one NVIDIA A800 80GB GPU. The low-cost RECAP risks in Table~\ref{tab:supp-probe-cost} use fewer probes without changing sample-level risks.}
\label{tab:runtime-summary}
\end{table}

\begin{table*}[t]
\centering
{\footnotesize
\setlength{\tabcolsep}{3pt}
\begin{tabular}{llrrrrrr}
\toprule
Model & Data & Full calls & Full Acc@80 & Full H-FPR & Staged calls & Staged Acc@80 & Staged H-FPR \\
\midrule
Qwen3-VL-8B & VSR & 4.8 & 85.3 & 16.9 & 3.1 & 85.9 & 18.7 \\
Qwen3-VL-8B & What'sUp & 5.0 & 85.1 & 20.4 & 3.2 & 82.6 & 24.1 \\
InternVL3.5-8B & VSR & 4.8 & 66.5 & 80.0 & 3.1 & 65.0 & 80.8 \\
InternVL3.5-8B & What'sUp & 5.0 & 51.1 & 70.6 & 3.2 & 51.0 & 70.7 \\
LLaVA-1.5-7B & VSR & 4.8 & 70.2 & 54.6 & 3.1 & 68.3 & 60.1 \\
LLaVA-1.5-7B & What'sUp & 5.0 & 58.7 & 57.9 & 3.2 & 56.2 & 59.8 \\
\bottomrule
\end{tabular}
}
\caption{Confidence-first staged deployment simulation. The policy first accepts the lowest-risk 60\% of samples under confidence, then runs RECAP-Img only on the remaining 40\% and accepts enough relation-ranked samples to reach 80\% total coverage. Calls/sample are expected likelihood calls; all metrics are computed on the final accepted set.}
\label{tab:staged-cost}
\end{table*}

\section{Direct-Answer Bias}

The main paper reports aggregate yes rates and direct hallucination false-positive rates. Here we apply a stricter control: a pure YesPenalty assigns the same rejection risk to every direct ``yes'' answer and therefore has no ranking ability within a fixed-answer subset. Table~\ref{tab:direct-answer-stratified} evaluates relation evidence after conditioning on the direct answer. RECAP-Img improves ranking in both direct-answer subsets on all three What'sUp models; on LLaVA-1.5-7B, Err-AUC rises from 73.4 to 85.2 within direct-yes answers and from 65.2 to 85.4 within direct-no answers. RECAP therefore exploits variation in relation evidence rather than merely imposing a global yes-answer penalty.

\begin{table*}[t]
\centering
{\footnotesize
\setlength{\tabcolsep}{4pt}
\begin{tabular}{lllrrrr}
\toprule
Model & Data & Direct answer & Risk & $n$ & Acc@80$\uparrow$ & Err-AUC$\uparrow$ \\
\midrule
Qwen3-VL-8B & VSR & yes & Confidence & 666 & 88.4 & 79.8 \\
 &  &  & RECAP-Img & 666 & 86.3 & 77.1 \\
 &  & no & Confidence & 583 & 82.9 & 67.9 \\
 &  &  & RECAP-Img & 583 & 83.9 & 70.5 \\
\midrule
Qwen3-VL-8B & What'sUp & yes & Confidence & 1459 & 65.0 & 91.6 \\
 &  &  & RECAP-Img & 1459 & 65.8 & 93.2 \\
 &  & no & Confidence & 1821 & 98.7 & 66.9 \\
 &  &  & RECAP-Img & 1821 & 99.1 & 80.1 \\
\midrule
InternVL3.5-8B & VSR & yes & Confidence & 1155 & 62.8 & 66.5 \\
 &  &  & RECAP-Img & 1155 & 63.9 & 72.0 \\
 &  & no & Confidence & 94 & 86.8 & 61.9 \\
 &  &  & RECAP-Img & 94 & 86.8 & 57.2 \\
\midrule
InternVL3.5-8B & What'sUp & yes & Confidence & 2744 & 36.7 & 83.5 \\
 &  &  & RECAP-Img & 2744 & 36.9 & 90.6 \\
 &  & no & Confidence & 536 & 99.3 & 67.2 \\
 &  &  & RECAP-Img & 536 & 99.8 & 73.1 \\
\midrule
LLaVA-1.5-7B & VSR & yes & Confidence & 990 & 63.0 & 60.6 \\
 &  &  & RECAP-Img & 990 & 67.0 & 72.1 \\
 &  & no & Confidence & 259 & 76.9 & 59.4 \\
 &  &  & RECAP-Img & 259 & 79.3 & 64.2 \\
\midrule
LLaVA-1.5-7B & What'sUp & yes & Confidence & 2482 & 36.3 & 73.4 \\
 &  &  & RECAP-Img & 2482 & 39.4 & 85.2 \\
 &  & no & Confidence & 798 & 97.0 & 65.2 \\
 &  &  & RECAP-Img & 798 & 98.6 & 85.4 \\
\bottomrule
\end{tabular}
}
\caption{Direct-answer-stratified selective prediction. RECAP-Img is evaluated within fixed direct-answer subsets, where a pure YesPenalty has no ranking ability.}
\label{tab:direct-answer-stratified}
\end{table*}

The corresponding What'sUp error directions rule out a simple class-rejection trade-off. Qwen's RECAP-Selector lowers FPR@80 from 23.6 to 19.4 while FNR@80 changes from 2.4 to 2.8. RECAP-Img lowers FPR@80 by 12.2 and 17.7 points for InternVL3.5-8B and LLaVA-1.5-7B; FNR@80 is 0.5 for InternVL3.5-8B and changes from 1.8 to 4.0 for LLaVA-1.5-7B. Thus the gains do not merely exchange false positives for false negatives.

Because H-FPR conditions on accepted negative statements, we also audit gold-yes and gold-no coverage under the same global 80\% threshold. Across the 12 model--dataset--label cells, 10 change by at most 2.1 points from Confidence to the deployed policy; the maximum change is 4.3 points on Qwen VSR, where yes/no coverage becomes more balanced (84.3/75.3 to 80.5/79.6). No class-specific threshold is fitted, so the gains do not reflect acceptance collapsing onto one label.

\begin{table}[t]
\centering
{\scriptsize
\setlength{\tabcolsep}{2.5pt}
\begin{tabular}{llrrrrr}
\toprule
Model & Data & 1\% & 2\% & 5\% & 10\% & 20\% \\
\midrule
Qwen3-VL-8B & VSR & 0 & 5 & 50 & 60 & 95 \\
 & What'sUp & 45 & 65 & 100 & 100 & 100 \\
InternVL3.5-8B & VSR & 0 & 0 & 0 & 0 & 0 \\
 & What'sUp & 0 & 0 & 0 & 0 & 0 \\
LLaVA-1.5-7B & VSR & 0 & 0 & 0 & 0 & 0 \\
 & What'sUp & 0 & 0 & 0 & 0 & 0 \\
\bottomrule
\end{tabular}
}
\caption{RECAP-Selector eligibility frequency over 20 group-disjoint splits as the labeled calibration fraction varies. Smaller budgets are nested within each primary 20\% calibration split; zero selects RECAP-Img. Values are percentages.}
\label{tab:calibration-size}
\end{table}

The gate's qualitative decision is stable with limited labels. InternVL3.5-8B and LLaVA-1.5-7B select evidence-only RECAP in every split at every budget. Qwen requires more evidence to certify confidence as complementary: selection becomes unanimous from 5\% on What'sUp and reaches 19/20 splits at 20\% on VSR, exactly matching the primary protocol. Threshold realization also stabilizes with sample size: held-out coverage standard deviation falls from 5.4--13.8 points at 1\% calibration to 1.4--3.0 points at 20\%. Small calibration sets can therefore preserve the mode regime, but reliable coverage control benefits from the full pre-specified budget.

\section{Equal-Budget Supervised Controls}

Probe-Logit is the direct supervised alternative to RECAP's hand-designed aggregation. It receives the same direct and image-probe margins before RECAP aggregation, together with relation-family indicators, and fits an L2-regularized logistic error model on each 20\% group-disjoint calibration split. Table~\ref{tab:supervised-combiner} reports held-out means over the same 20 seeds used in the main paper. Probe-Logit wins decisively on Qwen/What'sUp, revealing a strong exploitable in-domain bias. Fixed RECAP-Img is better on Acc@80 and H-FPR@80 in the other five settings and improves all four metrics on InternVL/VSR and both LLaVA datasets. On InternVL/What'sUp the fixed and learned risks have nearly identical selective utility, while Probe-Logit has 1.1 points higher Hall-AUC. Thus the structured score is not universally superior to supervised fitting; its advantage is strongest in lower-label or confidence-misaligned regimes where a low-capacity semantic prior avoids fitting dataset-specific answer patterns.

\begin{table*}[t]
\centering
{\footnotesize
\setlength{\tabcolsep}{3pt}
\begin{tabular}{llrrrrrr}
\toprule
Model & Data & \multicolumn{3}{c}{Probe-Logit} & \multicolumn{3}{c}{RECAP-Img} \\
\cmidrule(lr){3-5}\cmidrule(lr){6-8}
 & & Acc@80 & H-FPR & Hall-AUC & Acc@80 & H-FPR & Hall-AUC \\
\midrule
Qwen3-VL-8B & VSR & 81.1 & 18.4 & 65.9 & 85.0 & 17.4 & 68.7 \\
 & What'sUp & 95.0 & 5.0 & 98.2 & 85.1 & 20.4 & 79.2 \\
InternVL3.5-8B & VSR & 65.3 & 80.9 & 71.0 & 66.7 & 80.2 & 74.8 \\
 & What'sUp & 51.1 & 70.7 & 95.3 & 51.2 & 70.6 & 94.2 \\
LLaVA-1.5-7B & VSR & 68.7 & 58.8 & 73.5 & 69.9 & 55.2 & 77.3 \\
 & What'sUp & 56.9 & 58.9 & 89.3 & 58.6 & 58.0 & 91.0 \\
\bottomrule
\end{tabular}
}
\caption{Raw-probe supervised control under the same 20\% group-disjoint labeled calibration splits. Probe-Logit fits a model- and dataset-specific error separator; RECAP-Img keeps its formula fixed and uses labels only to set the threshold. Entries are held-out means in percent.}
\label{tab:supervised-combiner}
\end{table*}

As a deliberately richer upper bound, Logit-Hall additionally receives confidence, VCD, and RECAP diagnostic risks. It reaches Hall-AUC 90.5/98.4 for Qwen on VSR/What'sUp, 77.0/95.8 for InternVL, and 80.8/92.5 for LLaVA. This fusion shows what representative labels can add after feature engineering; it is an extension rather than an explanation for the fixed score's gains.

\section{Robustness and Scale Stress Tests}

\subsection{Full-Pool VSR Fallback}
\label{sec:full-vsr-fallback}

RECAP can be deployed without discarding relations outside its validated graph. For each of 20 image-disjoint splits, graph-covered statements use the calibration-selected RECAP mode and the other 946 statements use direct-answer confidence. Each branch is mapped through its calibration empirical CDF before one pooled calibration threshold targets 80\% coverage; no held-out label, outcome, or relation-specific threshold is used. Table~\ref{tab:full-vsr-fallback} evaluates this policy on all 2,195 VSR statements. The hybrid improves H-FPR and both ranking metrics for every model, while preserving Qwen accuracy and raising InternVL/LLaVA accuracy. Thus graph coverage controls where structured evidence is available, not whether the auditor can operate on the full benchmark.

\begin{table*}[t]
\centering
{\small
\setlength{\tabcolsep}{4pt}
\begin{tabular}{lrrrrrrrr}
\toprule
& \multicolumn{4}{c}{Confidence} & \multicolumn{4}{c}{RECAP + confidence fallback} \\
\cmidrule(lr){2-5}\cmidrule(lr){6-9}
Model & Acc@80 & H-FPR & Err-AUC & Hall-AUC & Acc@80 & H-FPR & Err-AUC & Hall-AUC \\
\midrule
Qwen3-VL-8B & 83.3 & 21.5 & 73.6 & 68.4 & 83.3 & \textbf{20.1} & \textbf{74.2} & \textbf{70.7} \\
InternVL3.5-8B & 62.1 & 95.0 & 63.3 & 61.8 & \textbf{64.7} & \textbf{86.9} & \textbf{69.4} & \textbf{69.8} \\
LLaVA-1.5-7B & 63.1 & 80.0 & 59.1 & 54.1 & \textbf{66.3} & \textbf{70.3} & \textbf{66.9} & \textbf{67.5} \\
\bottomrule
\end{tabular}
}
\caption{Full-pool VSR deployment over 20 image-disjoint splits. Graph-covered statements use RECAP; uncovered statements use confidence. Entries are held-out means in percent over all 2,195 statements.}
\label{tab:full-vsr-fallback}
\end{table*}

\subsection{External GSR-Bench Transfer}

Table~\ref{tab:supp-gsrbench-results} reports every pre-specified mode on the image-disjoint GSR-Bench subset. The evaluation uses COCO-Spatial-Two and GQA-Spatial-Two after removing every source-qualified image ID shared with VSR or controlled What'sUp, and provides RECAP with only RGB and the binary relation query. The dagger marks the mode selected from primary-domain calibration before inspecting GSR-Bench outcomes; the other rows prevent target-domain variant switching from being hidden.

\begin{table*}[t]
\centering
{\small
\setlength{\tabcolsep}{4pt}
\begin{tabular}{llrrrr}
\toprule
Model & Risk & Acc@80$\uparrow$ & H-FPR@80$\downarrow$ & Err-AUC$\uparrow$ & Hall-AUC$\uparrow$ \\
\midrule
Qwen3-VL-8B & Confidence & 91.5 & 2.5 & 77.2 & 81.0 \\
 & RECAP-Img & 93.0 & 0.5 & 84.9 & 96.5 \\
 & RECAP-Cal & 90.8 & 0.2 & 75.9 & 97.2 \\
 & RECAP-Selector$^\dagger$ & 91.2 & 0.6 & 79.3 & 94.9 \\
\midrule
InternVL3.5-8B & Confidence & 67.3 & 47.1 & 58.6 & 54.1 \\
 & RECAP-Img$^\dagger$ & 75.8 & 23.1 & 78.3 & 91.8 \\
 & RECAP-Cal & 75.3 & 24.0 & 73.7 & 91.9 \\
 & RECAP-Selector & 72.0 & 35.3 & 72.6 & 81.7 \\
\midrule
LLaVA-1.5-7B & Confidence & 56.8 & 94.6 & 67.0 & 64.2 \\
 & RECAP-Img$^\dagger$ & 66.1 & 76.9 & 88.3 & 93.0 \\
 & RECAP-Cal & 65.2 & 77.5 & 83.1 & 88.9 \\
 & RECAP-Selector & 60.9 & 89.5 & 81.3 & 80.6 \\
\bottomrule
\end{tabular}
}
\caption{External validation at common target-ranked 80\% coverage on the image-disjoint GSR-Bench subset ($1{,}438$ statements). $^\dagger$ marks the deployment mode selected on primary-domain calibration; RECAP-Cal remains a non-selected mechanism control. All entries are percentages.}
\label{tab:supp-gsrbench-results}
\end{table*}

Table~\ref{tab:strict-gsr-transfer} removes the remaining target-ranking convenience: for each of 20 source splits, both empirical CDFs and the nominal 80\% threshold are estimated from pooled VSR+What'sUp calibration groups and applied unchanged to GSR-Bench. No GSR score quantile or label is used. Realized coverage shifts across domains, so target calibration remains necessary when exact coverage is contractual. Under separately transferred thresholds, the source-selected image-risk mode lowers H-FPR and raises selective accuracy for InternVL and LLaVA; Qwen's selector reaches 0.8\% H-FPR at 74.6\% coverage. Because coverage differs between confidence and RECAP, these rows are deployment outcomes rather than equal-coverage comparisons.

\begin{table*}[t]
\centering
{\footnotesize
\setlength{\tabcolsep}{4pt}
\begin{tabular}{llrrrrrr}
\toprule
& & \multicolumn{3}{c}{Transferred Confidence} & \multicolumn{3}{c}{Transferred source-selected RECAP} \\
\cmidrule(lr){3-5}\cmidrule(lr){6-8}
Model & RECAP mode & Coverage & Accuracy & H-FPR & Coverage & Accuracy & H-FPR \\
\midrule
Qwen3-VL-8B & Selector & 61.0$\pm$3.8 & 94.1$\pm$0.6 & 1.5$\pm$0.1 & 74.6$\pm$3.3 & 92.6$\pm$0.5 & 0.8$\pm$0.2 \\
InternVL3.5-8B & Img & 44.7$\pm$2.3 & 71.9$\pm$0.3 & 52.8$\pm$0.8 & 86.0$\pm$0.7 & 73.6$\pm$0.4 & 29.9$\pm$1.0 \\
LLaVA-1.5-7B & Img & 85.6$\pm$1.0 & 55.7$\pm$0.2 & 93.0$\pm$0.2 & 62.0$\pm$1.6 & 77.8$\pm$1.1 & 59.4$\pm$2.4 \\
\bottomrule
\end{tabular}
}
\caption{Strict source-to-GSR deployment over 20 source calibration splits. Source CDFs, mode, and threshold are frozen before target evaluation; values are mean$\pm$standard deviation in percent. Realized coverage is reported because no target quantile is used.}
\label{tab:strict-gsr-transfer}
\end{table*}

\subsection{Prompt Robustness}

We evaluate Qwen3-VL-8B under four templates fixed before evaluation: the canonical relation question and three wrappers that request image-based, careful, or visible-evidence-only answers. Each wrapper is shared by the direct question and every relation probe. The resulting per-relation image and text margins support evaluation of both the default RECAP-Img and optional RECAP-Cal without additional inference. We report the mean and population standard deviation over all four templates and all six pairwise Spearman correlations; no best-template result is selected.

\begin{table*}[t]
\centering
{\small
\setlength{\tabcolsep}{3pt}
\begin{tabular}{llrrrrrr}
\toprule
Dataset & Risk & Acc@80$\uparrow$ & H-FPR@80$\downarrow$ & Err-AUC$\uparrow$ & Hall-AUC$\uparrow$ & Mean $\rho\uparrow$ & Min $\rho\uparrow$ \\
\midrule
VSR & Confidence & 84.3$\pm$0.9 & 19.8$\pm$4.7 & 74.4$\pm$0.6 & 70.7$\pm$0.9 & 0.913 & 0.866 \\
 & RECAP-Img & 84.8$\pm$0.7 & 17.9$\pm$3.1 & 74.1$\pm$0.7 & 70.9$\pm$2.6 & 0.912 & 0.886 \\
 & RECAP-Cal & 84.2$\pm$0.4 & 14.5$\pm$1.9 & 71.7$\pm$1.5 & 75.1$\pm$4.3 & 0.796 & 0.708 \\
\midrule
What'sUp & Confidence & 81.5$\pm$2.4 & 24.9$\pm$3.5 & 70.6$\pm$1.2 & 69.8$\pm$1.3 & 0.953 & 0.930 \\
 & RECAP-Img & 85.0$\pm$1.2 & 20.1$\pm$2.1 & 82.3$\pm$1.8 & 81.6$\pm$1.9 & 0.944 & 0.928 \\
 & RECAP-Cal & 87.2$\pm$2.1 & 16.6$\pm$3.2 & 85.2$\pm$3.4 & 84.9$\pm$3.8 & 0.900 & 0.856 \\
\bottomrule
\end{tabular}
}
\caption{Qwen3-VL-8B prompt robustness over four fixed templates. Values are mean $\pm$ population standard deviation across all templates; $\rho$ is pairwise Spearman risk-ranking correlation. No template is selected as best.}
\label{tab:prompt-robustness}
\end{table*}

The default RECAP-Img ranking is at least as stable as confidence: mean/minimum pairwise $\rho$ is 0.912/0.886 on VSR and 0.944/0.928 on What'sUp. Averaged over templates, it improves Acc@80 and H-FPR@80 on both datasets and improves both AUROCs on What'sUp. RECAP-Cal is more template-sensitive but retains its safety-oriented VSR and utility-oriented What'sUp gains. These results support the default image risk without prompt-wise selection and preserve the conclusion that text subtraction is optional rather than universally beneficial.

\subsection{Qwen3-VL-2B Scale Stress Test}

\begin{table}[t]
\centering
{\footnotesize
\setlength{\tabcolsep}{2pt}
\begin{tabular}{llrrr}
\toprule
Data & Risk & Acc@80 & H-FPR@80 & Hall-AUC \\
\midrule
VSR & Confidence & 74.0 & 44.5 & 63.8 \\
 & RECAP-Img & \textbf{76.2} & \textbf{30.1} & \textbf{75.3} \\
 & RECAP-Selector & 75.8 & 37.7 & 72.7 \\
\midrule
What'sUp & Confidence & 69.1 & 42.0 & 69.8 \\
 & RECAP-Img & \textbf{77.9} & \textbf{28.9} & \textbf{85.8} \\
 & RECAP-Selector & 74.7 & 34.1 & 81.9 \\
\bottomrule
\end{tabular}
}
\caption{Qwen3-VL-2B scale stress test under the fixed image-risk deployment modes. Bold marks the best risk per dataset; all entries are percentages.}
\label{tab:qwen2b-scale-supp}
\end{table}

The additional error-AUROC results follow the same pattern: RECAP-Img reaches 69.7 on VSR and 85.5 on What'sUp, versus 67.0 and 70.8 for confidence. Under the same 20 group-disjoint deployment splits as the main models, the eligibility test selects the confidence veto in 3/20 VSR splits and all 20 What'sUp splits. The resulting held-out coverage/accuracy/H-FPR is $80.2{\pm}1.9/76.1{\pm}0.9/31.4{\pm}3.9$ on VSR and $79.4{\pm}1.7/74.8{\pm}1.1/34.0{\pm}1.3$ on What'sUp. Thus both the fixed-risk gain and calibration-selected deployment extend to the smaller checkpoint.

\subsection{LLaVA-OneVision-7B Validation}

\begin{table*}[t]
\centering
{\small
\setlength{\tabcolsep}{4pt}
\begin{tabular}{llrrrr}
\toprule
Data & Risk & Acc@80$\uparrow$ & H-FPR@80$\downarrow$ & Err-AUC$\uparrow$ & Hall-AUC$\uparrow$ \\
\midrule
VSR & Confidence & 68.9 & 80.8 & 71.1 & 69.1 \\
 & VCD-Contrast & 73.8 & 62.1 & 77.6 & 78.2 \\
 & RECAP-Img & \textbf{74.9} & \textbf{58.5} & \textbf{80.1} & \textbf{80.9} \\
\midrule
What'sUp & Confidence & 56.5 & 62.0 & 68.6 & 68.0 \\
 & VCD-Contrast & 63.5 & 52.2 & 82.2 & 81.8 \\
 & RECAP-Img & \textbf{65.3} & \textbf{49.2} & \textbf{84.6} & \textbf{84.4} \\
\bottomrule
\end{tabular}
}
\caption{LLaVA-OneVision-7B held-out deployment over 20 group/image-disjoint calibration splits. Each calibration split fixes the 80\% coverage threshold; the eligibility test selects RECAP-Img in all splits. Entries are held-out means in percent.}
\label{tab:onevision-supp}
\end{table*}

The fixed-reference test selects RECAP-Img rather than the confidence veto in all 20 splits on both datasets. Held-out coverage/accuracy/H-FPR are $79.7\pm2.4$/$74.9\pm0.9$/$58.5\pm1.9$ on VSR and $80.1\pm2.2$/$65.3\pm0.7$/$49.2\pm0.6$ on What'sUp. In a separate pooled common-coverage paired bootstrap, gains over Confidence are positive for all metrics: on VSR, $+6.3$ Acc, $+22.5$ H-FPR reduction, $+9.3$ Err-AUC, and $+11.8$ Hall-AUC; on What'sUp, the gains are $+8.8$, $+13.0$, $+16.1$, and $+16.5$. All corresponding 95\% intervals exclude zero. This newer checkpoint reproduces the confidence-misaligned regime rather than merely inheriting the LLaVA-1.5 result.

\section{Failure Regimes and Cross-Model Auditing}

We organize recurring self-audit failures into evidence regimes and test whether a separate model can reduce correlated errors. These are aggregate diagnostics rather than claims of representative sample-level prevalence.

\subsection{Evidence Boundaries and Correlated Probe Errors}

RECAP is a self-auditing procedure: the same frozen model produces the direct answer and follow-up probes. Inspection reveals six recurring evidence regimes: object-recognition errors weaken both claim and anti evidence; binding errors attach evidence to the wrong pair; viewpoint or depth ambiguity narrows the claim--anti gap; jointly high evidence indicates relation inconsistency; jointly low evidence indicates insufficient visual support; and correlated answer--probe evidence supports the same wrong relation twice. These diagnostic categories localize where abstention suffices and where stronger perception, binding, or an external auditor is needed.

\begin{figure}[t]
\centering
\includegraphics[width=0.86\columnwidth]{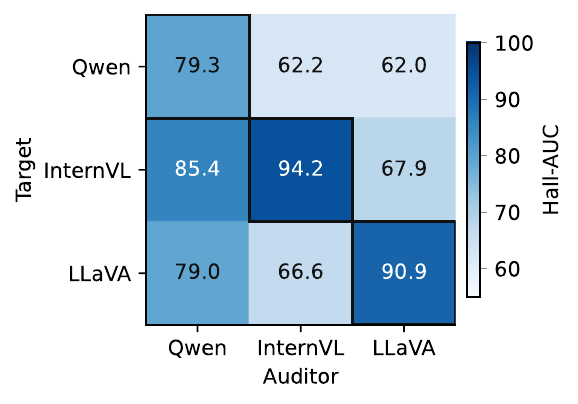}
\caption{Cross-model auditor check on What'sUp. The target model's direct answer and label are fixed; only RECAP-Img evidence comes from the auditor model. Entries show hallucination AUROC, and outlined cells mark self-auditing.}
\label{fig:cross-auditor}
\end{figure}

Figure~\ref{fig:cross-auditor} tests whether an external auditor can reduce model-coupled evidence errors. For each target model, we keep its direct answer and hallucination label fixed, but rank samples using RECAP-Img evidence from another model on the same image--query pair. Self-auditing remains strongest overall, showing that probe evidence is model-specific. Qwen nevertheless supplies useful external evidence for InternVL and LLaVA, reaching 85.4 and 79.0 hallucination AUROC. Cross-auditing therefore remains a boundary analysis rather than a headline deployment mode.

\section{Coverage and Relation-Family Analysis}

\subsection{Coverage Curves}

\begin{figure*}[t]
\centering
\includegraphics[width=0.92\textwidth]{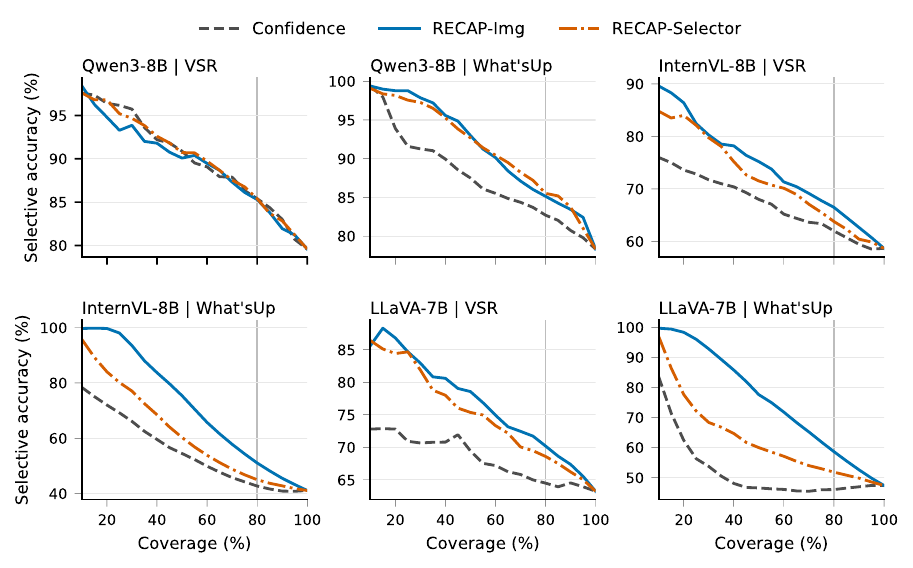}
\vspace{5pt}
\includegraphics[width=0.92\textwidth]{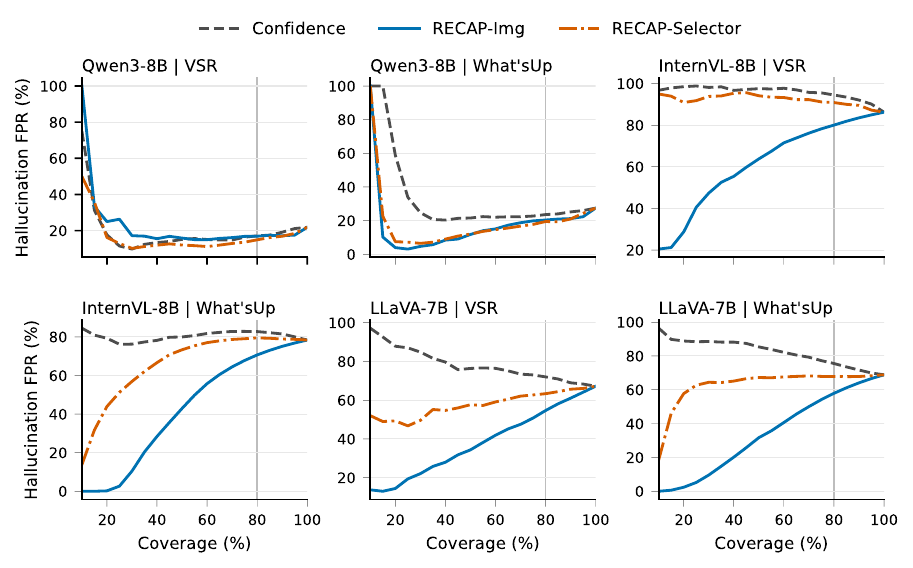}
\caption{Coverage curves for selective accuracy (top) and hallucination false-positive rate (bottom). Vertical lines mark the reported 80\% operating point; all curves use fixed risks. Image evidence is strongest when confidence is poorly aligned, while RECAP-Selector exposes the cost of retaining an unreliable confidence veto.}
\label{fig:supp-coverage-frontier}
\end{figure*}

Figure~\ref{fig:supp-coverage-frontier} shows the full risk--coverage behavior behind the fixed 80\% operating point. The same qualitative ordering persists at 70\%, 80\%, and 90\% coverage, with the largest separation when confidence is poorly aligned with relation grounding.

\begin{table}[t]
\centering
{\small
\setlength{\tabcolsep}{4pt}
\begin{tabular}{llrr}
\toprule
Model & Data & Confidence & Policy \\
\midrule
Qwen3-8B & VSR & 9.3 & 9.3 \\
 & What'sUp & 11.5 & 8.1 \\
InternVL-8B & VSR & 31.9 & 24.9 \\
 & What'sUp & 43.0 & 25.8 \\
LLaVA-7B & VSR & 30.2 & 22.7 \\
 & What'sUp & 44.8 & 22.3 \\
\bottomrule
\end{tabular}
}
\caption{Error AURC (lower is better). Deployed RECAP uses RECAP-Selector for Qwen and RECAP-Img otherwise, matching the pre-specified modes in the main paper. Values are percentages.}
\label{tab:supp-aurc}
\end{table}

Table~\ref{tab:supp-aurc} integrates selective error over the full coverage range. It confirms the fixed-coverage result: the policy is tied with confidence on Qwen/VSR and improves AURC in the other five settings, with the largest reductions for the confidence-misaligned models.

\subsection{Detailed Relation-Family Results}

Relation-family breakdowns test where the fixed constraint graph is reliable. All rows use the same pre-defined graph and scoring rules, so variation across families reveals its reliability profile. Cleanly exclusive pairs such as left/right, vertical, and inside/contains should benefit most, whereas depth is expected to be less stable because perspective, scale, and occlusion make both the claim and anti-relation probes noisier.

The leave-one-family-out test selects the deployment mode and target-80\% threshold using only the other families. It improves transferred H-FPR for the two confidence-misaligned models across all three held-out families. Qwen transfers well on left/right, is nearly tied on vertical, and favors confidence on depth, matching the reliability analysis below.

\paragraph{Family analysis.}
The compact breakdown preserves the same conclusion as the full per-risk tables. Topology on VSR and left/right on What'sUp provide the cleanest contradiction structure: RECAP-Img raises hallucination AUROC from 59.1 to 89.0 for LLaVA-1.5-7B topology and from 55.5 to 97.1 for LLaVA-1.5-7B left/right. Vertical relations also benefit across models, especially when confidence is yes-biased. Depth is the main boundary case: Qwen3-VL-8B remains better served by confidence, while RECAP improves depth ranking for InternVL3.5-8B and LLaVA-1.5-7B. Thus contradiction-aware evidence is strongest when the anti-relation is visually exclusive, and the same diagnostics identify relations that need image-adaptive or specialized probes.

\paragraph{Depth reliability.}
Table~\ref{tab:supp-depth-reliability} tests whether depth-edge reliability is measurable. For Qwen3-VL-8B, fixed depth edges over-commit on ambiguous examples, while a simple reliability gate falls back toward confidence. Reliable-depth subsets are substantially cleaner than ambiguous-depth subsets: fixed graph H-FPR is 10.1 vs. 34.8 for Qwen3-VL-8B, 87.6 vs. 99.6 for InternVL3.5-8B, and 66.0 vs. 90.2 for LLaVA-1.5-7B. The separation supports image-adaptive rather than globally hard depth edges.

\begin{table}[t]
\centering
{\footnotesize
\setlength{\tabcolsep}{2pt}
\begin{tabular}{llrr}
\toprule
Model & Variant & H-FPR@80 & Hall-AUC \\
\midrule
Qwen3-8B & Confidence & 16.5 & 76.3 \\
 & Fixed graph & 27.3 & 62.1 \\
 & Reliability gate & 17.6 & 74.1 \\
InternVL-8B & Confidence & 99.3 & 74.1 \\
 & Fixed graph & 94.0 & 90.1 \\
 & Reliability gate & 94.0 & 81.0 \\
LLaVA-7B & Confidence & 87.7 & 42.5 \\
 & Fixed graph & 74.7 & 79.6 \\
 & Reliability gate & 74.7 & 73.1 \\
\bottomrule
\end{tabular}
}
\caption{All-depth reliability results on controlled What'sUp. The gate is set from the calibration median of $|E_{\mathrm{pair}}|$; entries are percentages.}
\label{tab:supp-depth-reliability}
\end{table}

\raggedbottom
\section{Graph Extension and Multi-Choice Diagnostics}

The following experiments answer two distinct extension questions. The binary selective-audit analysis asks whether option-derived edges improve risk ranking for fixed yes/no answers; the four-way analysis asks whether those edges improve candidate relation selection. Keeping these tasks separate prevents gains in one from being read as gains in the other.

For binary selective auditing, each controlled option group proposes directed relation edges, and calibration retains an edge only when the gold relation wins at least 55\% of comparisons with a positive mean margin gap. The resulting 19--22 edges include 14--17 outside the manual graph. Relative to the manual graph, validation raises held-out Hall-AUC from 76.7 to 99.4 for Qwen, 85.7 to 93.5 for InternVL, and 77.5 to 87.4 for LLaVA. This supports extension when candidates are explicit, not unrestricted open-vocabulary induction.

Table~\ref{tab:supp-choice-score-comparison} asks the separate four-way selection question. Claim lift helps Qwen and InternVL, while fixed contradiction/support competition further raises InternVL accuracy from 76.3 to 85.0 and improves LLaVA's top-2 recovery. Auto-Validated edges match the image-margin Qwen result and provide smaller or mixed gains for the other models, so their strong binary-audit improvement does not imply uniform multi-choice improvement. They therefore remain an extension rather than a headline replacement.

Table~\ref{tab:supp-graph-sensitivity} isolates graph competition within the RECAP-Cal channel. Claim--anti competition recovers ranking lost by its claim-only counterpart; support edges matter less here because only \textit{on}$\rightarrow$\textit{above} is active. These rows therefore complement, rather than duplicate, the image-channel mechanism ablation in the main paper.

Held-out RECAP-Cal weight calibration raises Qwen Hall-AUC from 79.5 to 84.4, changes InternVL from 95.0 to 95.3, and leaves LLaVA unchanged at 87.8. We therefore keep fixed headline weights and treat adaptive edge reliability as an extension.

Among groups with a wrong top-1 relation, the gold relation is ranked second in 84.9\% of Qwen3-VL-8B cases (53 errors), 78.4\% of InternVL3.5-8B cases (287), and 60.8\% of LLaVA-1.5-7B cases (357). This is correction potential only: RECAP audits or ranks relations rather than rewriting deployed answers.

A separate full-pool grouped bootstrap quantifies the direct image-margin baseline for four-way selection. For Qwen3-VL-8B, Acc, MRR, and Top2 are 93.5$\pm$1.7, 96.6$\pm$0.9, and 99.0$\pm$0.7; the corresponding values are 65.0$\pm$3.2, 81.1$\pm$1.7, and 92.4$\pm$1.8 for InternVL3.5-8B, and 56.5$\pm$3.5, 75.2$\pm$2.0, and 82.9$\pm$2.6 for LLaVA-1.5-7B. These intervals characterize the unmodified candidate score rather than the held-out graph variants in Table~\ref{tab:supp-choice-score-comparison}. Depth accuracy has wider 95\% interval half-widths (5.6, 6.9, and 6.4 points), consistent with the smaller, visually harder family and the reliability analysis.

\begin{table}[H]
\centering
{\footnotesize
\setlength{\tabcolsep}{1.5pt}
\begin{tabular}{llrrr}
\toprule
Model & Candidate score & Accuracy & MRR & Top-2 recall \\
\midrule
Qwen3-8B & Image margin & 93.7 & 96.7 & 99.2 \\
 & Claim lift & 95.5 & 97.5 & 98.5 \\
 & Fixed RECAP graph & 95.7 & 97.7 & 99.1 \\
 & Auto-Validated graph & 93.7 & 96.7 & 99.2 \\
\midrule
InternVL-8B & Image margin & 64.9 & 81.0 & 92.0 \\
 & Claim lift & 76.3 & 87.2 & 94.6 \\
 & Fixed RECAP graph & 85.0 & 91.9 & 97.1 \\
 & Auto-Validated graph & 70.3 & 83.7 & 92.1 \\
\midrule
LLaVA-7B & Image margin & 56.3 & 75.0 & 81.7 \\
 & Claim lift & 56.5 & 75.3 & 84.8 \\
 & Fixed RECAP graph & 56.3 & 76.3 & 89.9 \\
 & Auto-Validated graph & 57.7 & 76.5 & 87.0 \\
\bottomrule
\end{tabular}
}
\caption{Held-out four-way relation selection on 646 controlled What'sUp groups. Fixed graph applies the pre-defined contradiction/support competition; Auto-Validated uses option-derived edges retained on disjoint calibration groups. All entries are percentages.}
\label{tab:supp-choice-score-comparison}
\end{table}

\begin{table}[H]
\centering
{\footnotesize
\setlength{\tabcolsep}{1.8pt}
\begin{tabular}{llrrrr}
\toprule
Model & Variant & Acc & H-FPR & Err-AUC & Hall-AUC \\
\midrule
Qwen3-8B & Confidence & 82.7 & 23.6 & 69.5 & 68.9 \\
 & Cal claim only & 76.6 & 32.8 & 52.0 & 49.9 \\
 & Cal claim+anti & 84.9 & 20.5 & 80.5 & 79.9 \\
InternVL-8B & Confidence & 42.7 & 82.8 & 67.0 & 66.9 \\
 & Cal claim only & 50.8 & 70.7 & 91.7 & 91.8 \\
 & Cal claim+anti & 51.2 & 70.6 & 95.0 & 95.1 \\
LLaVA-7B & Confidence & 46.0 & 75.6 & 52.2 & 51.0 \\
 & Cal claim only & 56.1 & 59.4 & 76.8 & 76.9 \\
 & Cal claim+anti & 57.5 & 58.6 & 87.1 & 87.6 \\
\bottomrule
\end{tabular}
}
\caption{Graph-edge sensitivity on controlled What'sUp using the RECAP-Cal channel $x=\Delta$. Removing anti/support edges weakens relation ranking, while claim--anti competition recovers most of the gain. Acc and H-FPR use 80\% coverage; all entries are percentages.}
\label{tab:supp-graph-sensitivity}
\end{table}

\end{document}